\documentclass{article}

\usepackage[preprint]{neurips_2026}

\usepackage[utf8]{inputenc} % allow utf-8 input
\usepackage[T1]{fontenc}    % use 8-bit T1 fonts
\usepackage{hyperref}       % hyperlinks
\usepackage{url}            % simple URL typesetting
\usepackage{booktabs}       % professional-quality tables
\usepackage{amsfonts}       % blackboard math symbols
\usepackage{nicefrac}       % compact symbols for 1/2, etc.
\usepackage{microtype}      % microtypography
\usepackage{xcolor}         % colors

\usepackage{amsmath}   % 提供数学环境
\usepackage{amssymb}   % 提供 \mathbb{R} 符号
\usepackage{bm}        % 提供 \bm{} 使希腊字母（如 \mu, \Sigma）变粗
\usepackage{multirow}  % 自己添加
\usepackage{graphicx}  % 自己添加

\title{Towards Physically Consistent 4D Scene Reconstruction for Closed-loop Autonomous Driving Simulation}

\author{%
  \textbf{Bowyn Tan}$^{1,2}$ \quad
  \textbf{Yutong Xie}$^{2}$ \quad
  \textbf{Bai Huang}$^{3}$ \quad
  \textbf{Fan Luo}$^{1,2}$ \quad
  \textbf{Xiao Li}$^{2}$ \\[0.5ex]
  \textbf{Naizheng Wang}$^{2}$ \quad
  \textbf{Yang Guan}$^{1,}$\thanks{Corresponding authors.} \quad
  \textbf{Shengbo Eben Li}$^{1,}$\footnotemark[1] \\[1ex]
  $^1$Tsinghua University \quad $^2$Meituan \quad
  $^3$Central University of Finance and Economics
}

\begin{document}

\maketitle
\begin{abstract}
High-fidelity street scene reconstruction is pivotal for end-to-end autonomous driving simulation, where novel-view synthesis (NVS) and time-varying information modeling are two fundamental capabilities to facilitate closed-loop training. However, existing 3DGS methods and their 4D extensions fail to simultaneously achieve both. To bridge this gap, we establish an information-geometric diagnostic framework, revealing that this limitation stems from a credit assignment dilemma between spatial and temporal parameters. Specifically, the deterministic coupling between viewpoint and time in single-source observation creates a low-rank structure that induces massive null-space ambiguity between static view-dependent and dynamic time-varying components. Temporal information overshadows spatial cues, causing the estimation variance of spatial parameters to diverge. To address this issue, we propose Orthogonal Projected Gradient (OPG), a hierarchical training method designed to restore spatial identifiability. OPG prioritizes the integrity of spatial representations by securing them in an initial stage, then restricts temporal updates to the spatial null space, enabling proactive credit assignment. While OPG isolates temporal updates algebraically, Temporal Regularization Strategy is proposed to further refine the temporal solution space by imposing a smoothness constraint based on the physical prior of consistent appearance evolution, ensuring that the reconstructed scene remains physically consistent in closed-loop simulation. Extensive experiments demonstrate that our method not only maintains stable NVS capabilities but also demonstrates superior performance in traditional observation-reproducing metrics, which indirectly reflect the capability of modeling temporal dynamics.
\end{abstract}

\section{Introduction}
\label{sec:inrtoduction}

End-to-end \textbf{autonomous driving (AD)} simulation necessitates high-fidelity scene reconstruction techniques that support closed-loop training~\cite{tian2025simscale, zhang2025carplanner, tian2025hgsim, hu2022model, gao2022sem2, li2024hydra}, imposing a rigorous requirement for both stable NVS capabilities and excellent time-varying information modeling ability. NVS requires the ego-agent may execute maneuvers that significantly diverge from the original data-collecting trajectory. As a powerful scene reconstruction method, 3D Gaussian Splatting(3DGS) and their 4D extensions have been widely adopted across various domains. Regarding street scenes, current Gaussian Splatting approaches can be split between purely spatial methods~\cite{zhou2024drivinggaussian, liu2025protocar, zhou2024hugs} that omitting temporal information, and temporal-aware methods~\cite{yan2024street, sun2025splatflow, song2025coda, ost2021neural, huang2024s3, chen2026periodic, li2025mtgs} that incorporate temporal dimension for time-varying modeling. However, we found that the latter methods suffer from severely poor NVS capabilities, despite achieving high reconstruction quality in observed-view reproducing. We gently argue that such methods are fundamentally insufficient for interactive simulation despite their remarkable visual quality metrics in observed-view reproducing.

\begin{figure}[t]
    \centering
    \includegraphics[width=1\linewidth]{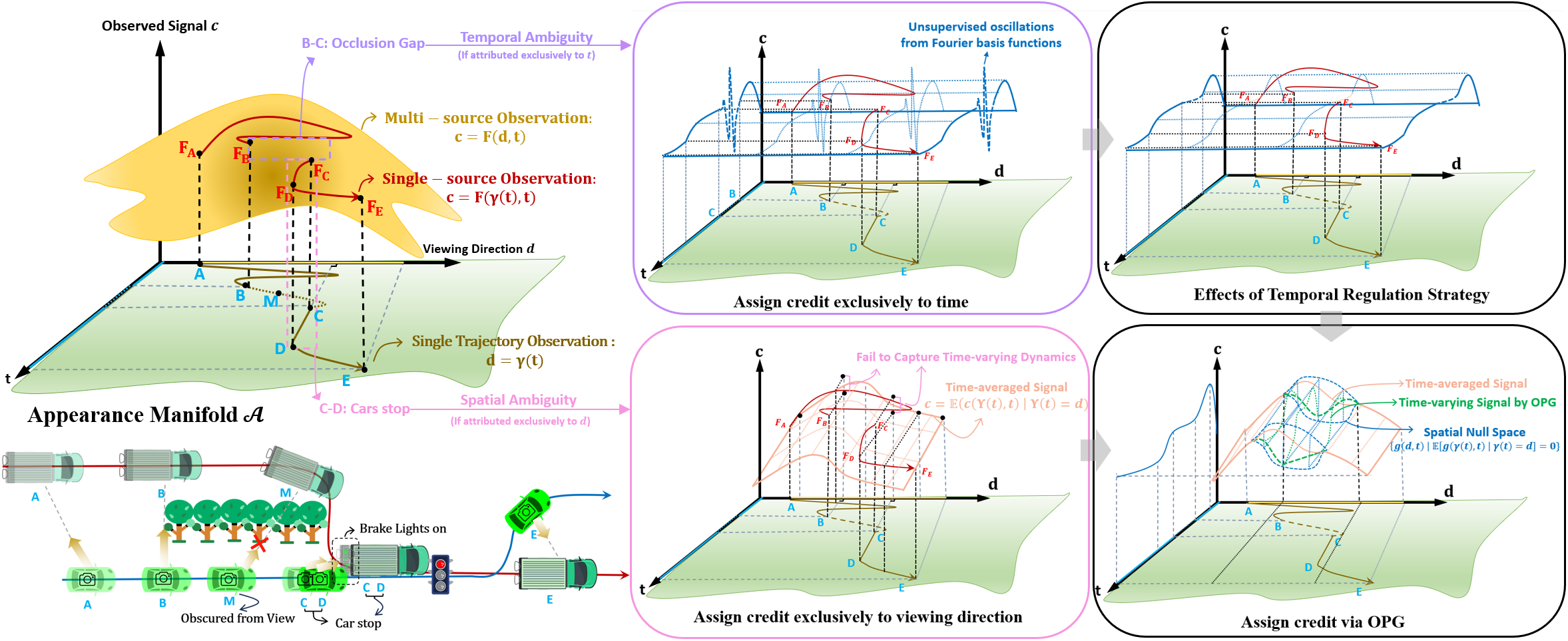}
     \caption{\textbf{Spatiotemporal Credit Assignment under SOF.} 4D reconstruction aims to recover the appearance surface $c=F(\mathbf{d}, t)$ (shorthand for $c_k$). While multi-source observation allows perfect surface solving, SOF restricts observations to a 1D trajectory $c=F(\gamma(t), t)$, necessitating proactive credit assignment via physical priors to infer the full surface manifold: \textbf{(1) Handling Occlusions (B-C):} During observation gaps, assigning credit to time leads to divergent estimation, whereas spatial assignment enables feature modeling by leveraging observations from other timestamps at the same viewpoint. \textbf{(2) Modeling Relative Statics (C-D):} When viewpoints are fixed, pure spatial assignment fails to capture temporal events like brake lights. With Temporal Regularization ensuring smooth evolution, OPG prioritizes spatial cues for stable extrapolation from other timestamps and model dynamics in spatial null space, achieving more accurate surface estimation.  }
    \label{fig:framework-diagnose}
\end{figure}

We are the first to clarify that this phenomenon stems from a credit assignment dilemma in the parameter space, which arises from the strong coupling of spatio-temporal information in AD scene reconstruction tasks: Unlike other scene reconstruction tasks, data collection in AD is restricted to a single driving trajectory with single-source observations, leading to a highly coupled spatiotemporal structure (Fig.~\ref{fig:framework-diagnose}), which we define as \textbf{Singular Observation Failure (SOF)}. We further demonstrate that temporal parameters hijacks spatial ones, leading to spatial underfitting. This occurs because the representational space of the latter is effectively a subspace of the former in SOF, allowing temporal parameters to overshadow the spatial ones.  In NVS scenarios, the exposure of inadequately modeled spatial feature results in a catastrophic mode collapse (demonstrated in Fig.~\ref{fig:core-demo}). Therefore, the price of introducing temporal parameters is the underfitting of spatial parameters caused by the overfitting of temporal parameters, resulting in drastic decline in NVS capability. 

To resolve the identifiability failure of spatial parameters caused by SOF, we propose a hierarchical training method via Orthogonal Projected Gradient(OPG). In the first stage, temporal parameters are frozen to ensure that time-invariant view-dependent spatial parameters are accurately learned without underfitting. In the second stage, we freeze the spatial parameters and optimize the temporal components by projecting and constraining them within the null space of the spatial parameters. This approach effectively alleviates spatial estimation issues caused by temporal information hijacking and ensures stable NVS performance. Furthermore, since the temporal dynamics are modeled within the spatial null space, the model’s temporal expressive power is fully preserved.

 To further ensure the reconstructured scene is physically consistent, we introduce Temporal Regularization Strategy grounded in the physical prior that object appearance evolves smoothly and consistently over time. This strategy restricts the expressiveness of temporal parameters and constrains their solution space by effectively excluding unphysical solutions, thereby effectively restricting the degrees of freedom within the temporal null space and alleviating overfitting. In conclusion, our contributions are three-fold:
\begin{figure}[t]
    \centering
    \includegraphics[width=1\linewidth]{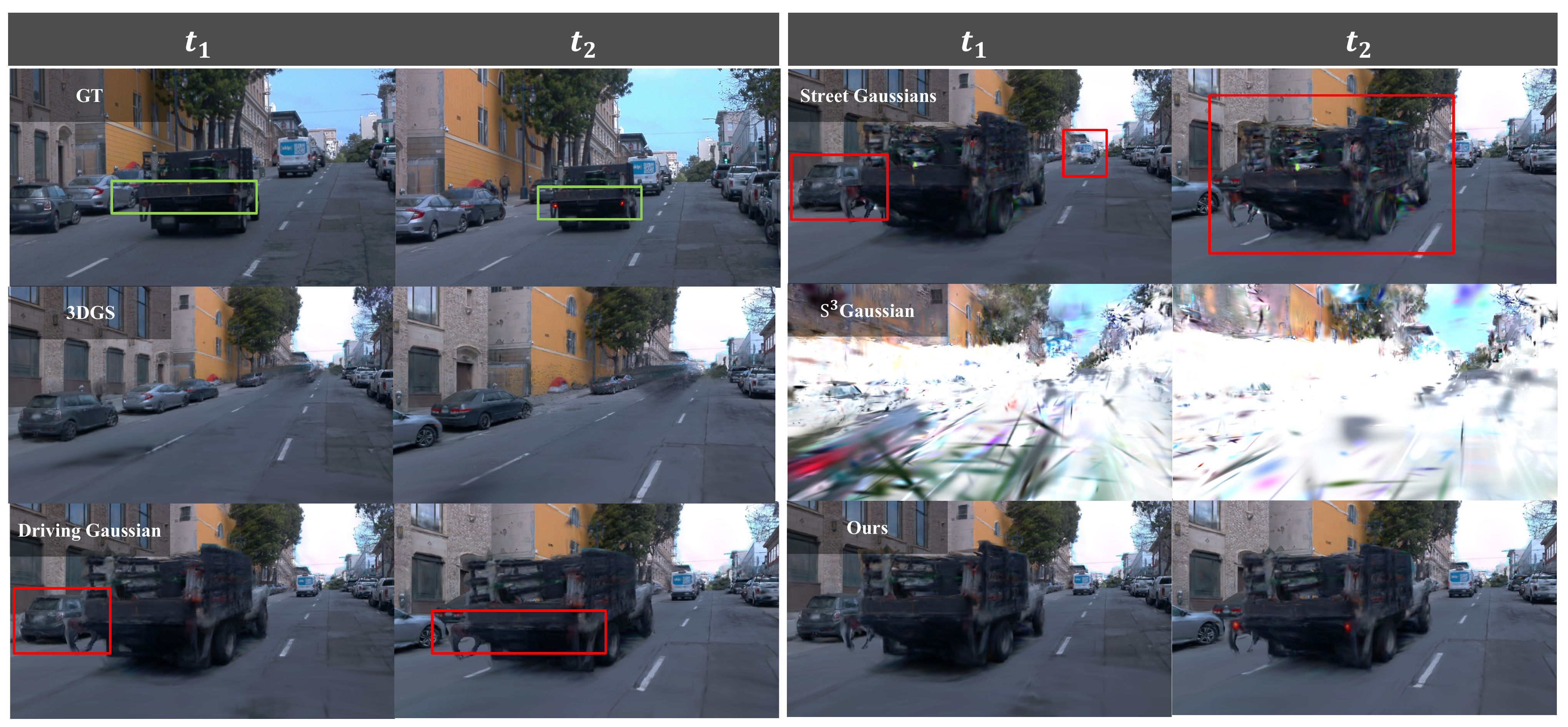}
    \caption{\textbf{Qualitative comparison of NVS capabilities.} 
    The Ground Truth (GT) captures a vehicle's taillights transitioning from \textit{off} ($t_1$) to \textit{on} ($t_2$). 
    Static \textbf{3DGS}~\cite{kerbl20233d} fails entirely on dynamic objects. 
    \textbf{DrivingGaussian}~\cite{zhou2024drivinggaussian} captures rigid motion but cannot model time-varying appearances, resulting in persistent dark lamps at $t_2$. 
    \textbf{StreetGaussians}~\cite{yan2024street} and \textbf{$S^3$Gaussian}~\cite{huang2024s3} attempt temporal modeling but suffer from severe morphological collapse in NVS. 
    Our method accurately synthesizes the taillight transition while maintaining robust NVS capabilities.}
    \label{fig:core-demo}
\end{figure}

\textbf{(1) Information-Geometric Diagnostic Framework for 4D Reconstruction:} We successfully attribute the mutual exclusivity of stable NVS and accurate temporal modeling in previous works to the ill-posed nature of SOF, which induces a credit assignment dilemma under spatio-temporal coupling, subsequently leading to the failure of parameter estimation.

\textbf{(2) Hierarchical training method via OPG:} We propose an OPG-based hierarchical training method, enabling proactive credit assignment through spatio-temporal decoupling. The initial stage is dedicated to ensuring that spatial parameters are accurately represented. In the subsequent stage, temporal optimization is further introduced and constrained to the unmodeled null space of these spatial parameters.

\textbf{(3) Temporal Regularization Strategy:} By introducing temporal total variation penalty, we leverage the physical prior of smooth appearance evolution to constrain the solution space of temporal parameters to physically consistent domain, ensuring robust NVS capabilities.

\section{Related works}

\paragraph{Gaussian Splatting methods} 
3DGS \cite{kerbl20233d, fu2024colmap}  has been widely adopted in static scene reconstruction tasks. For dynamic reconstruction, research has evolved into two main paradigms: (1) deformation-based 4DGS~\cite{wu20244d, zhang2025mega, fridovich2023k, gan2023v4d, lin2023high}, which uses implicit networks or feature grids to predict geometric and photometric offsets from a canonical state. However, its black-box nature hinders domain-prior injection, degrading robustness in under-observed conditions. (2) explicit spatiotemporal modeling~\cite{yang2024real, li2024spacetime, li2024spacetime}, which parameterizes time-varying attributes via explicit temporal coefficients per primitive. 

\paragraph{Scene reconstruction for driving simulation}
Current scene reconstruction for autonomous driving primarily follows two paradigms above. Within the \textbf{deformation-based lineage}, methods such as $S^3$Gaussian~\cite{huang2024s3}, SplatFlow~\cite{sun2025splatflow}, and CoDa-4DGS~\cite{song2025coda} employ self-supervised mechanisms---ranging from HexPlane decomposition to neural flow fields---to decouple dynamic components without 3D annotations. However, their implicit nature often causes physical motion to be misinterpreted as appearance fluctuations, leading to severe morphological breakdown during temporal extrapolation. Conversely, \textbf{explicit spatiotemporal modeling} enables better integration of physical priors. DrivingGaussian \cite{zhou2024drivinggaussian} ensures robust geometric completeness and superior view extrapolation under under-observed conditions by modeling entities as time-invariant static 3DGS objects. PVG~\cite{chen2026periodic} relies on transient primitives that vanish when unobserved, causing failure in NVS. StreetGaussians \cite{yan2024street} attempts to improve upon DrivingGaussian by introducing temporal mechanisms like 4D spherical harmonics, yet sacrifices the robust NVS ability inherent in DrivingGaussian. We address this defect through OPG and Temporal Regularization Strategy, restoring the balance between temporal modeling and NVS capabilities.

\section{Problem formulation}
\label{sec:problem_formulation}

In this section, we formalize the mathematical foundations of Gaussian Splatting and reformulate from the lens of information theory for a better quantitative evaluation of the problem's ill-posedness.

\subsection{Scene representation}
\begin{figure}[t]
    \centering
    \includegraphics[width=1\linewidth]{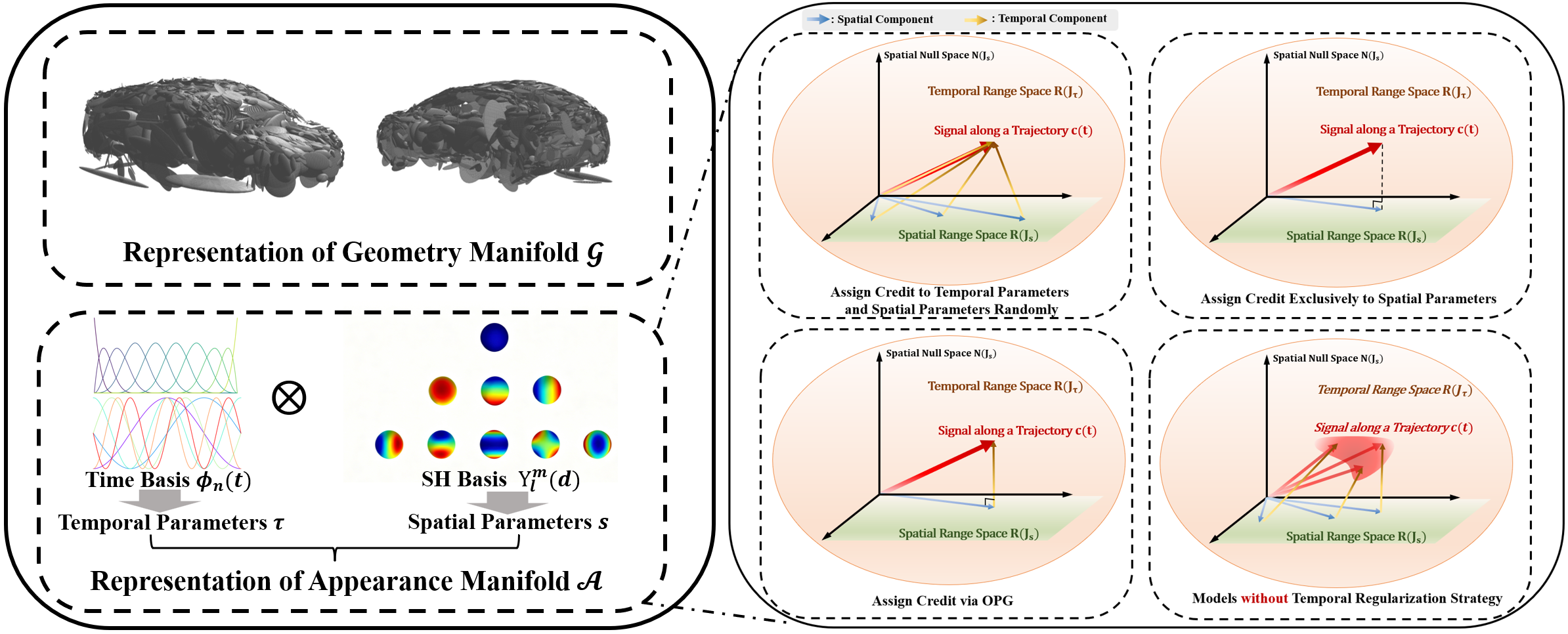}
    \caption{\textbf{Information-Geometric Diagnostic.} In SOF, the observed radiance $c(\gamma(t),t)$ can be modeled within the span of temporal bases $\mathcal{R}(\mathbf{J}_\tau)$. Unconstrained temporal component lead to solution ambiguity, while omitting them causes spatial underfitting. OPG ensures unique identifiability via proactive assignment. To handle observation gaps (cf. Fig.~\ref{fig:framework-diagnose} B-C) where unobserved intervals admit arbitrary functional values, expanding the solution $c(\gamma(t),t)$ into an under-determined manifold. Temporal Regularization Strategy constrains these intervals to enforce physical smoothness.}
    \label{fig:subframe}
\end{figure}

We represent a dynamic 4D scene as a structured manifold $\Theta$ composed of $M$ Gaussian primitives $\{\theta_k\}_{k=1}^M$. To support controllable and reactive simulation, we organize the Gaussian primitives into a structured scene graph $\Theta = \Theta_{\text{static}} \cup \Theta_{\text{dyn}}$, following recent dynamic reconstruction paradigms~\cite{zhou2024drivinggaussian, yan2024street,ost2021neural, miao2026evolsplat4d}, where $\Theta_{\text{static}}$ represents stationary environment elements, and $\Theta_{\text{dyn}}$ represents dynamic primitives capturing traffic participants. For each agent in $\Theta_{\text{dyn}}$, its global configuration is determined by a bounding box pose $\mathbf{T}$ which can be independently manipulated during simulation rollouts.  Each primitive admits a decomposition into a geometric component $\mathcal{G}$ and a spatiotemporal appearance component $\mathcal{A}$ shown in Fig.~\ref{fig:subframe}.

The Geometry Manifold $\mathcal{G} = \{ \mathbf{g}_k \}_{k=1}^M$ is modeled by a collection of Gaussian primitives. Each primitive $\mathbf{g}_k$ is defined by its center position $\bm{\mu}_k$, 3D covariance $\bm{\Sigma}_k$, and opacity $\alpha_k$, which collectively represent the physical structure of the scene. Crucially, we impose a \textbf{rigid-body prior} to dynamic objects by keeping these geometric parameters time-invariant within their local reference frames. This effectively prevents the non-physical shape distortions common in deformation-based 4DGS.

The Appearance Manifold $\mathcal{A} = \{ s_{lm}^{(k)},  \tau_n^{(k)} \}_{k=1}^M$ defines appearance features of each the Gaussian primitives, where $s_{lm}^{(k)}$ represents spatial parameters and $\tau_n^{(k)}$ represents temporal parameters. Following \cite{yang2024real, yan2024street}, \textbf{4D Spherindrical Harmonics (4DSH)} is employed to model the time-varying color $c_k(t, \mathbf{d})$, which represents the radiance emitted by the $k$-th Gaussian primitive at time $t$ along the viewing direction $\mathbf{d} \in \mathcal{S}^2$. It is formulated as a linear combination of composite basis functions:
\begin{equation}
\label{eq:4dsh_model}
    c_k(t, \mathbf{d}) = \sum_{l, m} s_{lm}^{(k)} \left( \sum_{n} \tau_n^{(k)} \phi_n(t) \right) Y_l^m(\mathbf{d}) = \sum_{n=0}^{N} \sum_{l=0}^{L} \sum_{m=-l}^{l} \alpha_{nlm}^{(k)} \underbrace{\left[ \phi_n(t) \cdot Y_l^m(\mathbf{d}) \right]}_{\mathcal{Z}_{nlm}(t, \mathbf{d})} 
\end{equation}
$\phi_n(t)$ represents the temporal basis functions, typically implemented as Fourier bases to capture time-varying information and $Y_l^m(\mathbf{d})$ represents the spherical harmonics (SH) used to capture spatial view-dependency (detailed in Appendix~\ref{sec:sh_details}). Their product, $\mathcal{Z}_{nlm}(t, \mathbf{d})$, constitutes the composite 4D spatiotemporal basis. The coefficients $s_{lm}$ govern the angular appearance variation, while $\tau_n$ govern the intrinsic evolution. It is worth noting that our discussion below focus mainly on appearance modeling. Sec.~\ref{sec:theoretical_framework} and Sec.~\ref{sec:opg_method} are premised on the assuming that the reconstruction of the geometric manifold is adequately precise.

\subsection{Differentiable rendering via spatiotemporal splatting}

To render the dynamic 4D scene into a 2D image plane at any given timestamp $t$, we formulate the process as a differentiable rasterization pipeline. We define the trainable parameters for each primitive as $\theta = \{ \theta_k \}_{k=1}^M = \{ \mathbf{g}_k, s_{lm}^{(k)}, \tau_n^{(k)} \}_{k=1}^M$. The rendering is performed under a query context $\mathbf{z} = (u, v, t, E, K)$ representing the pixel $(u, v)$, the timestamp $t$, the camera extrinsic $E$ and intrinsic $K$ matrices. The synthesized color $\mathbf{C}(\mathbf{z}; \theta)$ is computed by alpha-blending the $M$ gaussian primitives, sorted according to their depth (detailed in Appendix \ref{sec:3DGS_theory}):

\begin{equation}
\label{eq:diff_render}
\mathbf{C}(\mathbf{z}; \theta) = \sum_{k=1}^{M} \omega_k(\mathbf{g}_k, \mathbf{z}) \cdot c_k(s_{lm}^{(k)},  \tau_n^{(k)}, t, \mathbf{d}),
\end{equation}

where $c_k(s_{lm}^{(k)},  \tau_n^{(k)}, t, \mathbf{d})$ is the time-evolved and view-dependent color defined in Eq.~\ref{eq:4dsh_model} and the volumetric weights $\omega_k(\mathbf{g}_k, \mathbf{z})$ represent the visibility of each primitive detailed in Appendix~\ref{appendix:alpha_blending}. The equation computes the synthesized color via volumetric accumulation along the camera ray~\cite{yang2024real, kerbl20233d}. For brevity, we hereafter denote $\mathbf{C}(\mathbf{z}; \theta)$ simply as $\mathbf{C}(\theta)$.

\subsection{Quantifying Identifiability via Information Sensitivity}

To rigorously quantify the ill-posedness of the reconstruction, we adopt an information-geometric perspective~\cite{amari2012differential}. The local sensitivity of the observation to the parameters is captured by the Jacobian $\mathbf{J} = \frac{\partial \mathbf{C}(\theta)}{\partial \theta}$. Assuming Gaussian observation noise with variance $\sigma^2$, we define the Fisher Information Matrix (FIM)~\cite{fisher1925theory} (see Appendix \ref{sec:prob_formulation}) as:
\begin{equation}
\label{equation:Fish}
    \mathbf{F}(\theta) = \frac{1}{\sigma^2} \mathbf{J}^\top \mathbf{J}.
\end{equation}
Geometrically, $\mathbf{F}(\theta)$ acts as a precision metric in the parameter space. Its eigenvalues represent the amount of information available along different parameter directions. High eigenvalues correspond to directions where the observation is highly sensitive to parameter changes, ensuring stable estimation. Conversely, near-zero eigenvalues indicate an information-blind spot or a null-space, where parameters can drift arbitrarily without altering the rendered image.

The Cramér-Rao Bound (CRB)~\cite{rao1945information} provides the mathematical link between this geometric structure and estimation uncertainty:
\begin{equation}
\label{equation:CRB}
    \text{Cov}(\hat{\theta}) \succeq \mathbf{F}(\theta)^{-1} = \sigma^2 (\mathbf{J}^\top \mathbf{J})^{-1}.
\end{equation}

Eq.~\ref{equation:CRB} shows the spectrum of metric tensor $\mathbf{F}(\theta)$ governs parameter identifiability. Singularity in $\mathbf{F}(\theta)$ induces manifold collapse, creating a null-space where parameters drift without changing observations.

\section{Methods}
\label{sec:methods}

In this section, we formalize the mathematical foundations of our framework. We first provide a \textbf{information-geometry diagnostic framework} to analyze the degree of ill-posedness of Gaussian Splatting under various scenarios. We then introduce the \textbf{Orthogonal Projected Gradient (OPG)} method to mitigate the ill-posedness of the problem by enabling proactive credit assignment. Finally, we present reliable \textbf{Temporal Regularization Strategy} to further refine the temporal solution space, grounded in the physical prior of smooth temporal evolution in object appearance.

\subsection{Information-geometric diagnostic framework for 4D reconstruction}
\label{sec:theoretical_framework}

\paragraph{Fisher Information and Identifiability.} 
The precision of estimating the appearance manifold $\mathcal{A} = \{s_k, \tau_k\}$ is governed by the Fisher Information Matrix (FIM)~\cite{fisher1925theory}, which characterizes the sensitivity of the observation $\mathbf{C}$ to the parameters:
\begin{equation}
    \mathbf{F} = \frac{1}{\sigma^2} \mathbf{J}^\top \mathbf{J}, \quad \text{where} \quad \mathbf{J} = \left[ \mathbf{J}_s, \mathbf{J}_\tau \right] = \left[ \frac{\partial \mathbf{C}}{\partial s}, \frac{\partial \mathbf{C}}{\partial \tau} \right]
\end{equation}
The FIM can be partitioned into a block structure representing the auto-correlation and cross-correlation between spatial view-dependency $s$ and temporal evolution $\tau$:
\begin{equation}
    \mathbf{F} = \begin{bmatrix} 
    \mathbf{F}_{ss} & \mathbf{F}_{s\tau} \\ 
    \mathbf{F}_{\tau s} & \mathbf{F}_{\tau\tau} 
    \end{bmatrix} = \frac{1}{\sigma^2} \begin{bmatrix} 
    \mathbf{J}_s^\top \mathbf{J}_s & \mathbf{J}_s^\top \mathbf{J}_\tau \\ 
    \mathbf{J}_\tau^\top \mathbf{J}_s & \mathbf{J}_\tau^\top \mathbf{J}_\tau 
    \end{bmatrix}
\end{equation}

According to the Cramér-Rao Bound (CRB)~\ref{equation:CRB}, the covariance of unbiased estimators for spatial attributes $\hat{\mathbf{s}}$ and temporal evolution $\hat{\tau}$ is bounded by the inverse of their respective effective Fisher information matrices:
\begin{equation}
    \text{Cov}(\hat{\mathbf{s}}) \succeq \mathbf{S}_s^{-1}, \quad \text{Cov}(\hat{\tau}) \succeq \mathbf{S}_\tau^{-1}
\end{equation}
where $\mathbf{S}_s = \mathbf{F}_{ss} - \mathbf{F}_{s\tau} \mathbf{F}_{\tau\tau}^{-1} \mathbf{F}_{\tau s}$ and $\mathbf{S}_\tau = \mathbf{F}_{\tau\tau} - \mathbf{F}_{\tau s} \mathbf{F}_{ss}^{-1} \mathbf{F}_{s\tau}$ are the Schur complements within the joint FIM.

In an ideal full-rank sampling scenario like multi-source observation, the 4DSH basis functions $Z_{nl}^m(t, \mathbf{d}) = \phi_n(t) Y_l^m(\mathbf{d})$~\ref{eq:4dsh_model} are theoretically orthogonal (detailed in Appendix~\ref{sec:proof_org}), ensuring perfect identifiability. Consequently, the cross-modal Fisher information vanishes: $\mathbf{F}_{s\tau} = \frac{1}{\sigma^2} \mathbf{J}_s^\top \mathbf{J}_\tau = \mathbf{0}$. The covariance lower bounds for the spatial parameters $\hat{\mathbf{s}}$ and temporal parameters $\hat{\tau}$ are decoupled and finite:
\begin{equation}
    \text{Cov}(\hat{\mathbf{s}}) \succeq \mathbf{F}_{ss}^{-1}, \quad \text{Cov}(\hat{\tau}) \succeq \mathbf{F}_{\tau\tau}^{-1}.
\end{equation}
Under sufficient multi-source observations, this irreducible estimation uncertainty originates purely from the inherent observation noise $\sigma^2$ rather than parameter ambiguity, indicating a \textbf{well-posed} estimation problem.

\paragraph{Trajectory Degeneration and Orthogonality Collapse.}
In autonomous driving, however, the viewing direction $\mathbf{d}$ becomes a deterministic function of time $\mathbf{d}=\gamma(t)$ due to SOF. This functional dependence causes the basis functions to lose their orthogonality (detailed in Appendix~\ref{sec:proof_not_org}), triggering a \textbf{subspace inclusion} within the Jacobian space. Specifically, since the temporal basis $\{\phi_n(t)\}$ (e.g., Fourier series) is dense in $L^2([0, T])$, any trajectory-projected spatial signal can be approximated by the temporal manifold with arbitrary precision: $Y_l^m(\mathbf{d}(t)) = \sum_{\tilde{n}} \tau_{\tilde{n}} \phi_{\tilde{n}}(t)$.

 Consequently, the range space of the spatial Jacobian $\mathbf{J}_s$ becomes a subset of the temporal Jacobian $\mathbf{J}_\tau$ (proved in Appendix~\ref{sec:subspace}): $\mathcal{R}(\mathbf{J}_s) \subseteq \mathcal{R}(\mathbf{J}_\tau)$.

This subspace inclusion signifies a geometric collapse, rendering the Fisher metric tensor rank-deficient and inducing a non-trivial null-space. The spatial Jacobian $\mathbf{J}_s$ falls into the span of the temporal manifold $\mathbf{J}_\tau$, which implies the existence of a mapping $\mathbf{A}$ such that $\mathbf{J}_s \approx \mathbf{J}_\tau \mathbf{A}$. Substituting this into the Schur complement yields the effective Fisher information for $\mathbf{s}$:
\begin{equation}
    \mathbf{S}_s = \mathbf{F}_{ss} - \mathbf{F}_{s\tau} \mathbf{F}_{\tau\tau}^{-1} \mathbf{F}_{\tau s} \approx \mathbf{A}^\top \mathbf{F}_{\tau\tau} \mathbf{A} - (\mathbf{A}^\top \mathbf{F}_{\tau\tau}) \mathbf{F}_{\tau\tau}^{-1} (\mathbf{F}_{\tau\tau} \mathbf{A}) = \mathbf{0}.
\end{equation}
The collapse of $\mathbf{S}_s \to \mathbf{0}$ indicates that the estimation variance of spatial attributes diverges:
\begin{equation}
    \text{Cov}(\hat{\mathbf{s}}) \succeq \mathbf{S}_s^{-1} \to \infty.
\end{equation}

This divergence reveals our core conclusion: Because of SOF, spatial parameters $\mathbf{s}$ become fundamentally unidentifiable. The model fails to distinguish view-dependent reflections from temporal dynamics, allowing $\mathbf{s}$ to drift arbitrarily without penalizing the training loss. Consequently, this unconstrained spatial drift directly triggers the catastrophic rendering collapse in NVS when the change of spatial view point enables the misestimated $\mathbf{s}$ to contribute meaningfully to the reconstruction.

\subsection{Identifiability restoration via Orthogonal Projected Gradients}
\label{sec:opg_method}

To resolve the identifiability failure and suppress the variance explosion, we propose the hierarchical training method via \textbf{Orthogonal Projected Gradient (OPG)}, which enables proactive credit assignment by enforcing a structural constraint on the parameter updates. By projecting temporal gradients into the null-space of $\mathbf{J}_s$, we guarantee that the temporal manifold only captures variations that cannot be explained by spatial view-dependency.

\paragraph{Null-space projection.} 
We define a training-time temporal projector $\mathbf{P}_s^\perp$ that maps any gradient vector onto the null-space of the spatial manifold: $\mathbf{P}_s^\perp = \mathbf{I} - \mathbf{J}_s \mathbf{J}_s^\dagger$, where $\mathbf{J}_s^\dagger = (\mathbf{J}_s^\top \mathbf{J}_s)^{-1} \mathbf{J}_s^\top$ is the Moore-Penrose pseudoinverse. Instead of using the raw temporal Jacobian $\mathbf{J}_\tau$, OPG utilizes a purified temporal operator $\tilde{\mathbf{J}}_\tau$, defined by projecting $\mathbf{J}_\tau$ onto the spatial null-space:
$\tilde{\mathbf{J}}_\tau = \mathbf{P}_s^\perp \mathbf{J}_\tau = (\mathbf{I} - \mathbf{J}_s \mathbf{J}_s^\dagger) \mathbf{J}_\tau$. The purified operator satisfies $\mathbf{J}_s^\top \tilde{\mathbf{J}}_\tau = 0$, ensuring that temporal updates $\delta \bm{\tau}$ are strictly orthogonal to the spatial manifold.

\textbf{Reconditioning the FIM.} From an information-geometric perspective, OPG reconfigures the joint FIM into a block-diagonal structure:
\begin{equation}
    F_{OPG}(\theta) = \frac{1}{\sigma^2} \begin{bmatrix} \mathbf{J}_s^\top \mathbf{J}_s & \mathbf{0} \\ \mathbf{0} & \tilde{\mathbf{J}}_\tau^\top \tilde{\mathbf{J}}_\tau \end{bmatrix}.
\end{equation}
Under the OPG scheme, the enforcement of $\mathbf{J}_s^\top \tilde{\mathbf{J}}_\tau = \mathbf{0}$ ensures that the cross-modal Fisher information vanishes. Consequently, the Schur complements for both manifolds simplify to their respective diagonal blocks, yielding the following CRB bounds (detailed in Appendix\ref{sec:appendix_opg_proof}):
\begin{equation}
    \text{Cov}(\hat{\mathbf{s}}) \succeq \sigma^2 (\mathbf{J}_s^\top \mathbf{J}_s)^{-1}, \quad \text{Cov}(\hat{\tau}) \succeq \sigma^2 (\tilde{\mathbf{J}}_\tau^\top \tilde{\mathbf{J}}_\tau)^{\dagger}.
\end{equation}
where $(\cdot)^\dagger$ denotes the Moore-Penrose pseudoinverse. OPG mitigates the ill-posedness of the problem

\textbf{Hierarchical training method.} In the first training stage, temporal parameters are frozen to ensure that time-invariant spatial parameters are accurately learned without underfitting, prioritizing the physical correctness of static geometry and view-dependent appearance. In the second stage, we freeze the spatial parameters and optimize the temporal components by projecting and constraining them within the null space of the spatial parameters. By restricting $\hat{\tau}$ to a purified subspace, OPG successfully prevents temporal parameters from absorbing spatial residuals, ensuring robust novel-view extrapolation while introducing time-varying information.

\subsection{Temporal Regularization Strategy}
\label{sec:tv}
We introduced Temporal Regularization Strategies to further mitigate the overfitting of temporal parameters. Due to the moving of ego-vehicle and dynamic objects, a specific gaussian primitive is often observed only within a narrow temporal window $\mathcal{T}_{obs} \subset [0, T]$. 
Based on the prior of smooth temporal evolution in object appearance, Temporal Regularization Strategy leverages local transient observations to plausibly extrapolate features for unobserved time periods. This constrains the solution space within a physically consistent domain, thereby enhancing NVS performance. Specially, we introduce \textbf{Temporal Total Variation (TV) penalty} as:
\begin{equation}
\Psi(\tau) = \sum_{k=1}^{M} \int_{\mathcal{T}} \left\| \nabla_t \left( \sum_{n} \tau_n^{(k)} \phi_n(t) \right) \right\| dt \label{eq:tv_loss},    \mathcal{L}_{total} = \mathcal{L}_{traditional} + \lambda \Psi(\tau)
\end{equation}
Notably, an infinite $\lambda$ forces temporal parameters to be static, observations are dedicated to the estimation of low-dimensional spatial parameters, ensuring robust NVS while sacrificing temporal modeling. In contrast, a near-zero $\lambda$ removes constraints on temporal dynamics, allowing them to entangle with spatial information and causing NVS capabilities to deteriorate.

\section{Experiments}
\label{sec:experiments}
\subsection{The absense of suitable NVS metrics}
\label{sec:eval_gap}

It should be highlighted that while previous methods have adopted NVS metrics to validate their works, their NVS metrics are misleading and fails to reflect the performance when the vehicle deviates from the original data-collecting trajectory $\mathbf{d}=\gamma(t)$. Their so-called NVS is to test the temporal interpolation ablility, where test views never deviate from the single-source observation trajectory, thus SOF-related problems remain obscured. Nevertheless, constructing a universally accepted quantitative metric for it remains fundamentally challenging due to the strict absence of GT for unobserved novel views in real-world datasets. Therefore, to evaluate reconstruction models requires critically weighing these quantitative observed-view reproducing metrics against qualitative closed-loop stability. 

\subsection{Experimental setup}
\label{sec:setup}

\paragraph{Baseline models.}
We evaluate against SOTA paradigms to demonstrate our superior observation-reproducing performance and robust NVS capabilities: 
(1) \textbf{DrivingGaussian} \cite{zhou2024drivinggaussian}: By relying strictly on rigid-body transformations without temporal attributes, it ensures superior extrapolation albeit with lower training metrics. We re-implemented it within the StreetGaussians~\cite{yan2024street} codebase by replacing 4DSH with static 3DGS. 
(2) \textbf{StreetGaussians} \cite{yan2024street}: While its high degrees of freedom yield superior observation-view metrics, we demonstrate this is merely a manifestation of trajectory overfitting, resulting in catastrophic extrapolation collapse. 
(3) \textbf{$S^3$Gaussian} \cite{huang2024s3}: A deformation-based baseline without 3D box priors. Lacking explicit rigid constraints, it suffers severe morphological collapse during novel view extrapolation. Concurrent deformation-based methods (e.g., \textit{SplatFlow} \cite{sun2025splatflow}, \textit{CoDa-4DGS} \cite{song2025coda}) are excluded due to unavailable source codes, though we hypothesize they share the inherent extrapolation vulnerabilities of $S^3$Gaussian.

\textbf{Implementation details.} 
\label{sec:inplemetation_details}
We evaluate our method on the NOTR dataset, a representative subset of the Waymo Open Dataset \cite{sun2020scalability} curated by EmerNeRF \cite{yang2024emernerf}. All models are trained on a single NVIDIA A100 GPU at a resolution of $1600 \times 1066$ with three front cameras. Following the 3:1 interpolation protocol \cite{yan2024street}, we withhold every fourth frame for testing while using the remaining three for training, which is the so-called NVS metrics in previous methods. The integration of OPG scheme and TV regularization introduces negligible computational overhead, maintaining a training time almost identical to the original baseline. More inplementation details can be found in Appendix~\ref{appendix:inplementation}.

\subsection{Comparisons with the state-of-the-art}
\label{sec:comparison}

\begin{table}[h]
\centering
\small
\setlength{\tabcolsep}{3.5pt} % 统一列间距
\caption{Quantitative results on Waymo NOTR dataset. * denotes metrics within dynamic object regions. Our method achieves comparable performance on interpolation metrics while maintaining robust NVS capabilities.}
\label{tab:quantitative}
\begin{tabular}{l ccccc ccc}
\toprule
\multirow{2}{*}{\textbf{Method}} & \multicolumn{5}{c}{Waymo Dynamic32} & \multicolumn{3}{c}{Waymo Static32} \\
\cmidrule(lr){2-6} \cmidrule(lr){7-9}
 & \textbf{PSNR}$\uparrow$ & \textbf{SSIM}$\uparrow$ & \textbf{LPIPS}$\downarrow$ & \textbf{PSNR*}$\uparrow$ & \textbf{SSIM*}$\uparrow$ & \textbf{PSNR}$\uparrow$ & \textbf{SSIM}$\uparrow$ & \textbf{LPIPS}$\downarrow$ \\
\midrule
3DGS & 25.0657 & 0.8161 & 0.1899 & 19.3935 & 0.6110 & 26.0727 & 0.8129 & 0.1979 \\
DrivingGaussian & 26.6623& 0.8335 & 0.1675 & 23.8134& 0.7436 & 26.0859 & 0.8130 & 0.1977 \\
StreetGaussian & 26.7143& 0.8338& 0.1668& 24.3508& 0.7533& 26.0845& 0.8130 & 0.1978 \\
$S^3$Gaussian & 26.4839 & 0.8302 & 0.1881 & 22.5082 & 0.6313 & \textbf{26.6008} & \textbf{0.8207} & 0.1912 \\
\midrule
\textbf{Ours} & \textbf{26.9053}& \textbf{0.8375}& \textbf{0.1600}& \textbf{24.3672}& \textbf{0.7565}& 26.3079& 0.8170 & \textbf{0.1888}\\
\bottomrule
\end{tabular}
\end{table}

Table~\ref{tab:quantitative} presents the quantitative results on the Waymo NOTR dataset. We are surprised to find that our method happened to meet SOTA in several metrics, which may contribute to the introduce of OPG mitigates the overfitting to certain extent. We emphasize that maximizing traditional metrics is \textbf{not} our primary objective, its excellent performance demonstrates that our method possesses competitive capabilities in dynamic scene modeling. As visually corroborated in Fig.~\ref{fig:core-demo}, while prevailing methods achieve soaring interpolation metrics, they sacrifice structural integrity, leading to catastrophic artifacts during novel-view extrapolation. In contrast, by projecting temporal gradients into the spatial null-space, our framework substantially outperforms DrivingGaussian~\cite{zhou2024drivinggaussian} in open-loop tasks while preserving its superior extrapolation stability. This demonstrates a rare capability to achieve both intrinsic temporal modeling and physical consistency in AD simulation.

\subsection{Ablations and analysis}
\label{sec:ablations}

\begin{figure}[t]
    \centering
    \includegraphics[width=1\linewidth]{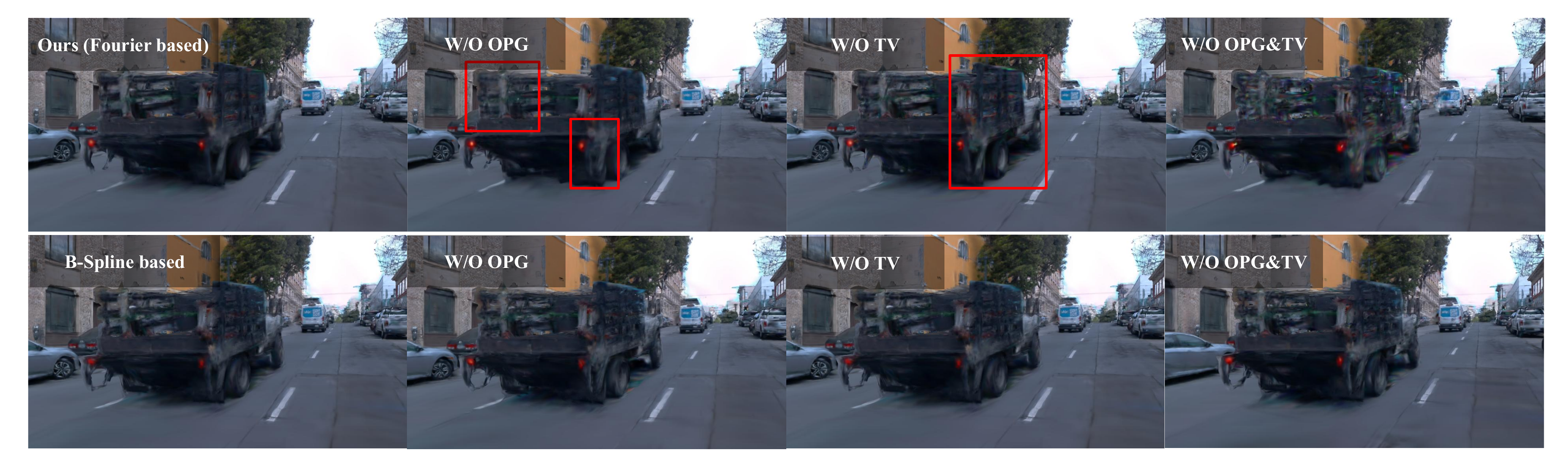}
    \caption{\textbf{Ablation study on temporal bases and model components.} Top and bottom rows show results using Fourier and B-spline bases, respectively. In each row, the leftmost image displays the full model, followed by its ablated versions.}
    \label{fig:ablation-study}
\end{figure}

\begin{table}[h]
\centering
\small 
\setlength{\tabcolsep}{3.5pt} 
\caption{Ablation study on Waymo NOTR dataset. * denotes metrics within dynamic object regions. ``OPG'' refers to orthogonal projected gradient; ``TV'' refers to temporal total variation. The default configuration uses Fourier series as the temporal basis.}
\label{tab:ablation_combined}
\begin{tabular}{l ccccc ccc}
\toprule
\multirow{2}{*}{\textbf{Configuration}} & \multicolumn{5}{c}{Waymo Dynamic32} & \multicolumn{3}{c}{Waymo Static32} \\
\cmidrule(lr){2-6} \cmidrule(lr){7-9}
 & \textbf{PSNR}$\uparrow$ & \textbf{SSIM}$\uparrow$ & \textbf{LPIPS}$\downarrow$ & \textbf{PSNR*}$\uparrow$ & \textbf{SSIM*}$\uparrow$ & \textbf{PSNR}$\uparrow$ & \textbf{SSIM}$\uparrow$ & \textbf{LPIPS}$\downarrow$ \\
\midrule
w/o OPG   & 26.7084 & 0.8339 & 0.1668 & 24.2388 & 0.7522 & 26.1116 & 0.8132 & 0.1975 \\
w/o TV & \textbf{26.9075} & 0.8373 & \textbf{0.1591} & \textbf{24.5692} & 0.7608 & \textbf{26.3335} & \textbf{0.8171} & \textbf{0.1886} \\
w/ B-spline Basis              & 26.7412 & 0.8335 & 0.1679 & 24.2157 & \textbf{0.7722} & 26.0817 & 0.8130 & 0.1977 \\
\midrule
\textbf{Full Framework (Ours)} & 26.9040 & \textbf{0.8374} & 0.1600 & 24.3667 & 0.7565 & 26.3079 & 0.8170 & 0.1888 \\
\bottomrule
\end{tabular}
\end{table}

We conduct ablation studies on the NOTR dataset to evaluate the core components of our framework: the Orthogonal Projected Gradient (OPG), the Temporal Total Variation (TV) penalty, and the choice of temporal basis. Quantitative and qualitative results are presented in Table~\ref{tab:ablation_combined} and Fig.~\ref{fig:ablation-study}, respectively.

\textbf{Orthogonal Projected Gradient (OPG).} Removing OPG degrades the quantitative reconstruction metrics and NVS capabilities(Table~\ref{tab:ablation_combined}). Without orthogonal decoupling, temporal parameters erroneously absorb viewpoint-dependent signals. Shown in Fig.~\ref{fig:ablation-study}, specular highlights become incorrectly baked into the object surface, blurring novel views and weakening overall extrapolation capability.

\textbf{Temporal TV regularization.} Removing TV slightly increases open-loop metrics (Table~\ref{tab:ablation_combined}), but degrades NVS capabilities (Fig.~\ref{fig:ablation-study}). The phenomenon of trade-off is consistent with our expectations.

\textbf{Fourier series vs. B-spline basis.} Since the 4DSH formulation imposes no strict constraints on the temporal basis $\phi_n(t)$, we explored B-splines as an alternative. When training without OPG or TV, B-splines are superior because their local support confines the influence of transient observations to specific time windows, acting as an implicit regularization against sequence-wide contamination. However, once OPG and TV restore algebraic identifiability,  high-frequency representational capacity of Fourier series becomes a decisive advantage, enabling higher-fidelity reconstructions(Table~\ref{tab:ablation_combined}).

\section{Conclusion}
\label{sec:conclusion}

In this paper, we introduced an information-geometric diagnose framework based on 
CRB to analyze the degree of ill-posedness of the reconstructiion problem. Our analysis mathematically demonstrates that the inherent limitations of single-source data-collecting trajectories lead to Singular Observation Failure (SOF). Under SOF, spatial parameters become entangled with temporal dynamics, making the spatial parameters unestimatable and resulting in the failure of NVS. To fundamentally resolve this degeneracy, we proposed OPG scheme. By projecting temporal gradients into the null space of spatial parameters, OPG enables proactive credit assignment. Furthermore, we integrated Temporal Regularization Strategy to further confine the extensive temporal solution space. Extensive experiments demonstrate that our approach not only maintains both robust NVS capabilitie and time-varying information modeling. For future work, we plan to extend this representation into a generative world model for closed-loop autonomous driving simulation.

\paragraph{Limitations}
\label{sec:limitations}
The effectiveness of our OPG method is predicated on high-quality geometric optimization, otherwise, artifacts like "floaters" may capture temporal dynamics, distorting novel views. We suggest that while TV loss serves as a general-purpose stabilizer for enhancing novel-view extrapolation, OPG specifically necessitates a clean geometric foundation to avoid amplifying artifacts. Additionally, our framework lacks the capacity to model non-rigid motion, such as pedestrians.

\bibliographystyle{unsrt}
\bibliography{references}

\begin{thebibliography}{10}

\bibitem{tian2025simscale}
Haochen Tian, Tianyu Li, Haochen Liu, Jiazhi Yang, Yihang Qiu, Guang Li, Junli
  Wang, Yinfeng Gao, Zhang Zhang, Liang Wang, et~al.
\newblock Simscale: Learning to drive via real-world simulation at scale.
\newblock {\em arXiv preprint arXiv:2511.23369}, 2025.

\bibitem{zhang2025carplanner}
Dongkun Zhang, Jiaming Liang, Ke~Guo, Sha Lu, Qi~Wang, Rong Xiong, Zhenwei
  Miao, and Yue Wang.
\newblock Carplanner: Consistent auto-regressive trajectory planning for
  large-scale reinforcement learning in autonomous driving.
\newblock In {\em Proceedings of the Computer Vision and Pattern Recognition
  Conference}, pages 17239--17248, 2025.

\bibitem{tian2025hgsim}
Yue Tian, Wenbo Chu, Wei Zhou, Xiaolin Tang, and Keqiang Li.
\newblock Hgsim: High-fidelity and generalizable simulation frame-work for
  autonomous driving scenes.
\newblock {\em Neurocomputing}, page 131784, 2025.

\bibitem{hu2022model}
Anthony Hu, Gianluca Corrado, Nicolas Griffiths, Zachary Murez, Corina Gurau,
  Hudson Yeo, Alex Kendall, Roberto Cipolla, and Jamie Shotton.
\newblock Model-based imitation learning for urban driving.
\newblock {\em Advances in Neural Information Processing Systems},
  35:20703--20716, 2022.

\bibitem{gao2022sem2}
Zeyu Gao, Yao Mu, Ruoyan Shen, Chen Chen, Yangang Ren, Jianyu Chen,
  Shengbo~Eben Li, Ping Luo, and Yanfeng Lu.
\newblock Sem2: Enhance sample efficiency and robustness of end-to-end urban
  autonomous driving via semantic masked world model.
\newblock In {\em Deep Reinforcement Learning Workshop NeurIPS 2022}, 2022.

\bibitem{li2024hydra}
Zhenxin Li, Kailin Li, Shihao Wang, Shiyi Lan, Zhiding Yu, Yishen Ji, Zhiqi Li,
  Ziyue Zhu, Jan Kautz, Zuxuan Wu, et~al.
\newblock Hydra-mdp: End-to-end multimodal planning with multi-target
  hydra-distillation.
\newblock {\em arXiv preprint arXiv:2406.06978}, 2024.

\bibitem{zhou2024drivinggaussian}
Xiaoyu Zhou, Zhiwei Lin, Xiaojun Shan, Yongtao Wang, Deqing Sun, and Ming-Hsuan
  Yang.
\newblock Drivinggaussian: Composite gaussian splatting for surrounding dynamic
  autonomous driving scenes.
\newblock In {\em Proceedings of the IEEE/CVF conference on computer vision and
  pattern recognition}, pages 21634--21643, 2024.

\bibitem{liu2025protocar}
Hongyuan Liu, Haochen Yu, Bochao Zou, Juntao Lyu, Qi~Mei, Jiansheng Chen, and
  Huimin Ma.
\newblock Protocar: Learning 3d vehicle prototypes from single-view and
  unconstrained driving scene images.
\newblock In {\em Proceedings of the AAAI Conference on Artificial
  Intelligence}, volume~39, pages 5460--5468, 2025.

\bibitem{zhou2024hugs}
Hongyu Zhou, Jiahao Shao, Lu~Xu, Dongfeng Bai, Weichao Qiu, Bingbing Liu, Yue
  Wang, Andreas Geiger, and Yiyi Liao.
\newblock Hugs: Holistic urban 3d scene understanding via gaussian splatting.
\newblock In {\em Proceedings of the IEEE/CVF Conference on Computer Vision and
  Pattern Recognition}, pages 21336--21345, 2024.

\bibitem{yan2024street}
Yunzhi Yan, Haotong Lin, Chenxu Zhou, Weijie Wang, Haiyang Sun, Kun Zhan,
  Xianpeng Lang, Xiaowei Zhou, and Sida Peng.
\newblock Street gaussians: Modeling dynamic urban scenes with gaussian
  splatting.
\newblock In {\em European Conference on Computer Vision}, pages 156--173.
  Springer, 2024.

\bibitem{sun2025splatflow}
Su~Sun, Cheng Zhao, Zhuoyang Sun, Yingjie~Victor Chen, and Mei Chen.
\newblock Splatflow: Self-supervised dynamic gaussian splatting in neural
  motion flow field for autonomous driving.
\newblock In {\em Proceedings of the Computer Vision and Pattern Recognition
  Conference}, pages 27487--27496, 2025.

\bibitem{song2025coda}
Rui Song, Chenwei Liang, Yan Xia, Walter Zimmer, Hu~Cao, Holger Caesar, Andreas
  Festag, and Alois Knoll.
\newblock Coda-4dgs: Dynamic gaussian splatting with context and deformation
  awareness for autonomous driving.
\newblock In {\em Proceedings of the IEEE/CVF International Conference on
  Computer Vision}, pages 28031--28041, 2025.

\bibitem{ost2021neural}
Julian Ost, Fahim Mannan, Nils Thuerey, Julian Knodt, and Felix Heide.
\newblock Neural scene graphs for dynamic scenes.
\newblock In {\em Proceedings of the IEEE/CVF Conference on Computer Vision and
  Pattern Recognition}, pages 2856--2865, 2021.

\bibitem{huang2024s3}
Nan Huang, Xiaobao Wei, Wenzhao Zheng, Pengju An, Ming Lu, Wei Zhan, Masayoshi
  Tomizuka, Kurt Keutzer, and Shanghang Zhang.
\newblock {S$^3$Gaussian}: Self-supervised street gaussians for autonomous
  driving.
\newblock {\em arXiv preprint arXiv:2405.20323}, 2024.

\bibitem{chen2026periodic}
Yurui Chen, Chun Gu, Junzhe Jiang, Xiatian Zhu, and Li~Zhang.
\newblock Periodic vibration gaussian: Dynamic urban scene reconstruction and
  real-time rendering.
\newblock {\em International Journal of Computer Vision}, 134(3):83, 2026.

\bibitem{li2025mtgs}
Tianyu Li, Yihang Qiu, Zhenhua Wu, Carl Lindstr{\"o}m, Peng Su, Matthias
  Nie{\ss}ner, and Hongyang Li.
\newblock Mtgs: Multi-traversal gaussian splatting.
\newblock {\em arXiv preprint arXiv:2503.12552}, 2025.

\bibitem{kerbl20233d}
Bernhard Kerbl, Georgios Kopanas, Thomas Leimk{\"u}hler, George Drettakis,
  et~al.
\newblock 3d gaussian splatting for real-time radiance field rendering.
\newblock {\em ACM Trans. Graph.}, 42(4):139--1, 2023.

\bibitem{fu2024colmap}
Yang Fu, Sifei Liu, Amey Kulkarni, Jan Kautz, Alexei~A Efros, and Xiaolong
  Wang.
\newblock Colmap-free 3d gaussian splatting.
\newblock In {\em Proceedings of the IEEE/CVF conference on computer vision and
  pattern recognition}, pages 20796--20805, 2024.

\bibitem{wu20244d}
Guanjun Wu, Taoran Yi, Jiemin Fang, Lingxi Xie, Xiaopeng Zhang, Wei Wei, Wenyu
  Liu, Qi~Tian, and Xinggang Wang.
\newblock 4d gaussian splatting for real-time dynamic scene rendering.
\newblock In {\em Proceedings of the IEEE/CVF conference on computer vision and
  pattern recognition}, pages 20310--20320, 2024.

\bibitem{zhang2025mega}
Xinjie Zhang, Zhening Liu, Yifan Zhang, Xingtong Ge, Dailan He, Tongda Xu, Yan
  Wang, Zehong Lin, Shuicheng Yan, and Jun Zhang.
\newblock Mega: Memory-efficient 4d gaussian splatting for dynamic scenes.
\newblock In {\em Proceedings of the IEEE/CVF International Conference on
  Computer Vision}, pages 27828--27838, 2025.

\bibitem{fridovich2023k}
Sara Fridovich-Keil, Giacomo Meanti, Frederik~Rahb{\ae}k Warburg, Benjamin
  Recht, and Angjoo Kanazawa.
\newblock K-planes: Explicit radiance fields in space, time, and appearance.
\newblock In {\em Proceedings of the IEEE/CVF conference on computer vision and
  pattern recognition}, pages 12479--12488, 2023.

\bibitem{gan2023v4d}
Wanshui Gan, Hongbin Xu, Yi~Huang, Shifeng Chen, and Naoto Yokoya.
\newblock V4d: Voxel for 4d novel view synthesis.
\newblock {\em IEEE Transactions on Visualization and Computer Graphics},
  30(2):1579--1591, 2023.

\bibitem{lin2023high}
Haotong Lin, Sida Peng, Zhen Xu, Tao Xie, Xingyi He, Hujun Bao, and Xiaowei
  Zhou.
\newblock High-fidelity and real-time novel view synthesis for dynamic scenes.
\newblock In {\em SIGGRAPH Asia 2023 Conference Papers}, pages 1--9, 2023.

\bibitem{yang2024real}
Zeyu Yang, Hongye Yang, Zijie Pan, and Li~Zhang.
\newblock Real-time photorealistic dynamic scene representation and rendering
  with {4D} gaussian splatting.
\newblock In {\em The Twelfth International Conference on Learning
  Representations (ICLR)}, 2024.

\bibitem{li2024spacetime}
Zhan Li, Zhang Chen, Zhong Li, and Yi~Xu.
\newblock Spacetime gaussian feature splatting for real-time dynamic view
  synthesis.
\newblock In {\em Proceedings of the IEEE/CVF Conference on Computer Vision and
  Pattern Recognition}, pages 8508--8520, 2024.

\bibitem{miao2026evolsplat4d}
Sheng Miao, Sijin Li, Pan Wang, Dongfeng Bai, Bingbing Liu, Yue Wang, Andreas
  Geiger, and Yiyi Liao.
\newblock Evolsplat4d: Efficient volume-based gaussian splatting for 4d urban
  scene synthesis.
\newblock {\em arXiv preprint arXiv:2601.15951}, 2026.

\bibitem{amari2012differential}
Shun-ichi Amari.
\newblock {\em Differential-geometrical methods in statistics}.
\newblock Springer Science \& Business Media, 2012.

\bibitem{fisher1925theory}
Ronald~Aylmer Fisher.
\newblock Theory of statistical estimation.
\newblock In {\em Mathematical proceedings of the Cambridge philosophical
  society}, volume~22, pages 700--725. Cambridge University Press, 1925.

\bibitem{rao1945information}
C~Radhakrishna Rao et~al.
\newblock Information and the accuracy attainable in the estimation of
  statistical parameters.
\newblock {\em Bull. Calcutta Math. Soc}, 37(3):81--91, 1945.

\bibitem{sun2020scalability}
Pei Sun, Henrik Kretzschmar, Xerxes Dotiwalla, Aurelien Chouard, Vijaysai
  Patnaik, Paul Tsui, James Guo, Yin Zhou, Yuning Chai, Benjamin Caine, et~al.
\newblock Scalability in perception for autonomous driving: Waymo open dataset.
\newblock In {\em Proceedings of the IEEE/CVF conference on computer vision and
  pattern recognition}, pages 2446--2454, 2020.

\bibitem{yang2024emernerf}
Jiawei Yang, Boris Ivanovic, Or~Litany, Xinshuo Weng, Seung~Wook Kim, Boyi Li,
  Tong Che, Danfei Xu, Sanja Fidler, Marco Pavone, and Yue Wang.
\newblock Emer{N}e{RF}: Emergent spatial-temporal scene decomposition via
  self-supervision.
\newblock In {\em International Conference on Learning Representations}, 2024.

\end{thebibliography}

\appendix
\label{Appendix}

\section{Detailed formulation of 3D Gaussian Splatting}
\label{sec:3DGS_theory}

This section provides a comprehensive mathematical description of the 3D Gaussian Splatting (3DGS) framework, extending the summary provided in Section 3 of the main text.

\subsection{Geometric parameterization of primitives}
As defined in Sec.~3.1, each Gaussian primitive $\theta_k$ is characterized by its geometric manifold $\mathbf{g}_k = \{ \bm{\mu}_k, \alpha_k, \bm{\Sigma}_k \}$. To ensure that the 3D covariance matrix $\bm{\Sigma}_k \in \mathbb{R}^{3 \times 3}$ remains positive semi-definite (PSD) during gradient-based optimization, it is decomposed into a scaling matrix $\mathbf{S}_k$ and a rotation matrix $\mathbf{R}_k$:
\begin{equation}
\bm{\Sigma}_k = \mathbf{R}_k \mathbf{S}_k \mathbf{S}_k^\top \mathbf{R}_k^\top
\end{equation}
where $\mathbf{S}_k = \text{diag}(s_x, s_y, s_z)$ is a diagonal matrix representing the scale along the three principal axes, and $\mathbf{R}_k$ is a rotation matrix derived from a normalized unit quaternion $\mathbf{q}_k = [r, x, y, z]$. For primitives in $\Theta_{\text{dyn}}$, these parameters are anchored in the local reference frame of the respective agent to satisfy the rigid-body prior.

\subsection{Differentiable rendering and alpha-blending}
\label{appendix:alpha_blending}
\paragraph{Projected 2D Covariance} 
To render the 3D primitives into a 2D image plane, the 3D covariance $\bm{\Sigma}_k$ is projected into camera coordinates. Given the camera extrinsic matrix $\mathbf{E}$ and the Jacobian $\mathbf{J}_{\text{proj}}$ of the affine approximation of the perspective projection, the projected 2D covariance $\bm{\Sigma}_k^{2d}$ is computed following the EWA Splatting framework:
\begin{equation}
\bm{\Sigma}_k^{2d} = \mathbf{J}_{\text{proj}} \mathbf{E} \bm{\Sigma}_k \mathbf{E}^\top \mathbf{J}_{\text{proj}}^\top
\end{equation}

The alpha-blending equation is:
\begin{equation}
\label{eq:diff_render1}
\mathbf{C}(\mathbf{z}; \theta) = \sum_{k=1}^{M} \omega_k(\mathbf{g}_k, \mathbf{z}) \cdot c_k(s_{lm}^{(k)},  \tau_n^{(k)}, t, \mathbf{d}),
\end{equation}

where $c_k(s_{lm}^{(k)},  \tau_n^{(k)}, t, \mathbf{d})$ is the time-evolved and view-dependent color. Equation~\ref{eq:diff_render1} computes the synthesized color via volumetric accumulation along the camera ray, where the volumetric weights $\omega_k(\mathbf{g}_k, \mathbf{z})$ represent the visibility of each primitive . The term $\omega_k(\mathbf{g}_k, \mathbf{z})$ is explicitly formulated as:

\begin{equation}
\omega_k(\mathbf{g}_k, \mathbf{z}) = p_k(u, v | \bm{\mu}_k^{2d}, \bm{\Sigma}_k^{2d}) \alpha_k \prod_{j=1}^{k-1} \left( 1 - p_j(u, v | \bm{\mu}_j^{2d}, \bm{\Sigma}_j^{2d}) \alpha_j \right),
\end{equation}

where $\alpha_k \in \mathbf{g}_k$ denotes the intrinsic opacity. The 2D probability density $p_k$ is computed by projecting the 3D Gaussian onto the image plane:
\begin{equation}
    p_k(u, v | \bm{\mu}_k^{2d}, \bm{\Sigma}_k^{2d}) = \exp \left( -\frac{1}{2} ([u, v]^\top - \bm{\mu}_k^{2d})^\top (\bm{\Sigma}_k^{2d})^{-1} ([u, v]^\top - \bm{\mu}_k^{2d}) \right),
\end{equation}
where the projected 2D mean $\bm{\mu}_k^{2d}$ and covariance $\bm{\Sigma}_k^{2d}$ are obtained as:
\begin{equation}
    \bm{\mu}_k^{2d} = \text{Proj}(\bm{\mu}_k | E, K)_{1:2}, \quad \bm{\Sigma}_k^{2d} = (\mathbf{J}_{proj} E \bm{\Sigma}_k E^\top \mathbf{J}_{proj}^\top)_{1:2,1:2},
\end{equation}
where $\text{Proj}( \cdot | E, K)$ denotes the transformation from the world space to the image space and $\mathbf{J}_{proj}$ is the Jacobian of the perspective projection.

\subsection{View-dependent appearance via spherical harmonics}

To model non-Lambertian effects such as specular reflections, the color $c_k$ of each Gaussian is parameterized as a function of the normalized viewing direction $\mathbf{d} \in S^2$. The normalized viewing direction $\mathbf{d} = (\theta, \phi)$ is defined as the unit vector pointing from the Gaussian center $\bm{\mu}$ to the camera optical center $\mathbf{o}$:
\begin{equation}
\mathbf{d} = \frac{\mathbf{o} - \bm{\mu}}{\|\mathbf{o} - \bm{\mu}\|_2}
\end{equation}

\section{Spatiotemporal modeling and orthogonality analysis}

\subsection{4D Spherindrical Harmonics (4DSH)}

To reconstruct dynamic scenes, the static color model is extended to the temporal domain $\mathcal{T} = [0, T]$. The time-varying appearance of a Gaussian primitive is defined on the product manifold $\mathcal{M} = \mathcal{T} \times S^2$. We represent the spatiotemporal color $c(t, \mathbf{d})$ using a Spherindrical Harmonics expansion:
\begin{equation}
c(t, \mathbf{d}) = \sum_{n=0}^{N} \sum_{l=0}^{L} \sum_{m=-l}^{l} \alpha_{nlm} \underbrace{\left[ \phi_n(t) \cdot Y_l^m(\mathbf{d}) \right]}_{\mathcal{Z}_{nlm}(t, \mathbf{d})}
\end{equation}
where $\phi_n(t)$ are temporal basis functions, $Y_l^m(\mathbf{d})$ are spatial SH bases, and $\mathcal{Z}_{nlm}(t, \mathbf{d})$ are the composite 4D spatiotemporal basis functions.

\subsection{Full-rank independent sampling (The decoupled ideal)}
\label{sec:proof_org}
To analyze the ill-posedness, we examine the continuous limit of the Jacobians. Let the spatiotemporal basis be $\mathcal{Z}_{nlm}(t, \mathbf{d}) = \phi_n(t) Y_l^m(\mathbf{d})$. In an ideal scenario where the viewing direction $\mathbf{d}$ and time $t$ are sampled independently over the full manifold $\mathcal{T} \times S^2$, the inner product can be expanded using \textbf{Fubini's Theorem}:
\begin{equation}
\begin{aligned}
\langle \mathcal{Z}_{nlm}, \mathcal{Z}_{n'l'm'} \rangle &= \int_{\mathcal{T}} \int_{S^2} \left[ \phi_n(t) Y_l^m(\mathbf{d}) \right] \cdot \left[ \phi_{n'}(t) Y_{l'}^{m'}(\mathbf{d}) \right] d\Omega dt \\
&= \underbrace{\left( \int_{\mathcal{T}} \phi_n(t) \phi_{n'}(t) dt \right)}_{\text{Temporal Integral}} \cdot \underbrace{\left( \int_{S^2} Y_l^m(\mathbf{d}) Y_{l'}^{m'}(\mathbf{d}) d\Omega \right)}_{\text{Spatial Integral}} \\
&= \delta_{nn'} \cdot \delta_{ll'} \cdot \delta_{mm'}
\end{aligned}
\end{equation}
where $\delta_{ij}$ is the Kronecker delta. This property ensures that temporal evolution and view-dependency are uniquely identifiable.

The separability of the double integral ensures that the temporal and spatial Jacobians are mutually orthogonal ($\mathbf{J}_\tau^T \mathbf{J}_s = \mathbf{0}$). This leads to a block-diagonal Gram matrix $\mathbf{G}$, allowing for perfect attribution of residuals.

\subsection{Proof of orthogonality collapse on 1D trajectories}
\label{sec:proof_not_org}

In autonomous driving (AD), the measure is restricted to a 1D trajectory $\mathbf{d}=\gamma(t)  $. The empirical inner product on $\gamma$ is:
\begin{equation}
\langle \mathcal{Z}_{nlm}, \mathcal{Z}_{n'l'm'} \rangle_{\gamma} = \int_{0}^{T} \phi_n(t) Y_l^m(\mathbf{d}(t)) \cdot \phi_{n'}(t) Y_{l'}^{m'}(\mathbf{d}(t)) \, dt \neq \delta_{nn'} \delta_{ll'} \delta_{mm'}
\end{equation}

\paragraph{Orthogonality collapse}
For a smooth 1D trajectory $\gamma(t)$ where $\mathbf{d}$ is a deterministic function of $t$, the basis functions $\mathcal{Z}_{nlm}$ lose their orthogonality. Specifically, the coherence $\rho = \langle \mathcal{Z}_{nlm}, \mathcal{Z}_{n'l'm'} \rangle_{\gamma} \neq 0$ for indices where temporal and spatial frequencies overlap.

\subsection{Proof of subspace inclusion on 1D trajectories}
\label{sec:subspace}

We consider the total pixel color $\mathbf{C}$ as defined by the aggregation of $K$ Gaussian primitives along a 1D trajectory $\mathbf{d}=\gamma(t)  $. Based on the provided formulation, the composite color and the $k$-th component's spatiotemporal expansion are given by:
\begin{equation}
\mathbf{C}(t) = \sum_{k=1}^K c_k(t, \mathbf{d}(t)) \cdot \omega_k, \quad c_k(t, \mathbf{d}) = \sum_{l,m} s_{lm}^{(k)} Y_l^m(\mathbf{d}) \cdot \sum_n \tau_n \phi_n(t)
\end{equation}
where $\omega_k$ are rendering weights independent of the color parameters $\{\tau, s\}$. To prove the subspace inclusion $\mathcal{R}(\mathbf{J}_s) \subseteq \mathcal{R}(\mathbf{J}_\tau)$, we derive the partial derivatives of the composite color $\mathbf{C}$ with respect to the temporal and spatial coefficients.

\paragraph{1. Temporal gradient formulation} Applying the chain rule to the total color $\mathbf{C}$ with respect to the $n$-th temporal coefficient $\tau_n$:
\begin{equation}
\begin{aligned}
g_{\tau, n}^{(k)}(t) = \frac{\partial \mathbf{C}}{\partial \tau_n^{(k)}} &=  \omega_k \frac{\partial c_k(t, \mathbf{d}(t))}{\partial \tau_n^{(k)}} \\
&=  \omega_k \left[ \left( \sum_{l,m} s_{lm}^{(k)} Y_l^m(\mathbf{d}(t)) \right) \phi_n(t) \right] \\
&= \phi_n(t) \cdot \underbrace{\left[  \omega_k \sum_{l,m} s_{lm}^{(k)} Y_l^m(\mathbf{d}(t)) \right]}_{B(t)}
\end{aligned}
\end{equation}
where $B(t)$ represents the total projected spatial contribution of all Gaussians at time $t$. Each temporal gradient component is a modulation of the basis $\phi_n(t)$ by the aggregate spatial signal $B(t)$.

\paragraph{2. Spatial gradient formulation} For a specific spatial coefficient $s_{lm}^{(k)}$ of the $k$-th Gaussian, the gradient of $\mathbf{C}$ is:
\begin{equation}
\begin{aligned}
g_{s, lm}^{(k)}(t) = \frac{\partial \mathbf{C}}{\partial s_{lm}^{(k)}} &= \omega_k \frac{\partial c_k(t, \mathbf{d}(t))}{\partial s_{lm}^{(k)}} \\
&= \omega_k \left[ Y_l^m(\mathbf{d}(t)) \cdot \sum_n \tau_n^{(k)} \phi_n(t) \right] \\
&= \omega_k Y_l^m(\mathbf{d}(t)) \cdot T(t)
\end{aligned}
\end{equation}
where $T(t) = \sum_n \tau_n^{(k)} \phi_n(t)$ is the shared temporal modulation function.

\paragraph{3. Proof of subspace inclusion} We seek a set of coefficients $\{\gamma_n\}$ such that the spatial gradient $g_{s, lm}^{(k)}(t)$ is a linear combination of the temporal gradients $\{g_{\tau, n}(t)\}$. This requires:
\begin{equation}
\begin{aligned}
\omega_k Y_l^m(\mathbf{d}(t)) T(t) &= \sum_{n} \gamma_n [ B(t) \phi_n(t) ] \\
\frac{\omega_k Y_l^m(\mathbf{d}(t)) T(t)}{B(t)} &= \sum_{n} \gamma_n \phi_n(t)
\end{aligned}
\end{equation}
Since $\mathbf{d}(t)$ is a continuous trajectory, the term $H(t) = \frac{\omega_k Y_l^m(\mathbf{d}(t)) T(t)}{B(t)}$ is a well-defined function in $L^2([0, T])$. Given that $\{\phi_n(t)\}$ is a complete Fourier or polynomial basis, there exists a unique representation of $H(t)$ in the span of $\{\phi_n\}$. Consequently, the spatial Jacobian $\mathbf{J}_s$ satisfies:
\begin{equation}
\mathcal{R}(\mathbf{J}_s) \subseteq \mathcal{R}(\mathbf{J}_\tau)
\end{equation}

\section{Variational analysis of geometry-appearance entanglement}

To rigorously characterize the indistinguishability between motion and appearance, we define the total variation of the observed color signal $\delta \mathbf{C}$ within a unified variational framework. For a scene represented by spatiotemporal primitives, the variation is a coupled manifestation of geometric and photometric manifolds:
\begin{equation}
\delta \mathbf{C} = \mathbf{J}_g \delta \mathbf{g} + \mathbf{J}_s \delta \mathbf{s} + \mathbf{J}_\tau \delta \mathbf{\tau}
\end{equation}
where $\mathbf{J}_g, \mathbf{J}_s, \mathbf{J}_\tau$ are the Jacobians with respect to the geometric ($\delta \mathbf{g}$), spatial appearance ($\delta \mathbf{s}$), and temporal appearance ($\delta \mathbf{\tau}$) variations, respectively.

\paragraph{Dimensionality reduction via rigid-body prior}
A fundamental distinction between primitive-based methods  and grid-based representations lies in the structural definition of $\mathbf{J}_g$. In autonomous driving scenarios, 3DGS-based frameworks typically reconstruct the scene within an ego-local coordinate system. By assuming the vehicle and its sensor suite as an \textbf{ideal rigid body}, the geometric manifold is constrained to a lower-dimensional subspace. This prior effectively anchors the point cloud, significantly reducing the rank of $\mathbf{J}_g$. Consequently, the \textbf{cross-null-space} $\mathcal{N}(\mathbf{J}_{total})$, where $\mathbf{J}_g \delta \mathbf{g} + \mathbf{J}_s \delta \mathbf{s} + \mathbf{J}_\tau \delta \mathbf{\tau} = \mathbf{0}$, is substantially constricted compared to global volumetric methods.

\paragraph{The HexPlane divergence}
In contrast, HexPlane-based paradigms operate without explicit rigid-body priors, modeling the scene as a continuous 4D volume $[X, Y, Z, T]$. This global representation treats every spatiotemporal voxel as an independent degree of freedom, leading to a high-dimensional geometric Jacobian $\mathbf{J}_g$ with massive null-space intersections. Without the structural anchoring of local primitives, HexPlane suffers from severe \textbf{Morphological Deconstruction}, where physical motion is erroneously interpreted as stochastic fluctuations within the feature planes, leading to non-physical distortions in novel views.

\section{Detailed expressions for spherical harmonics}
\label{sec:sh_details}

In our framework, the spatial view-dependency is modeled using real spherical harmonics $Y_l^m(\mathbf{d})$. Here, $\mathbf{d}$ is the \textbf{normalized viewing direction vector} pointing from the Gaussian center to the camera, represented in Cartesian coordinates as $\mathbf{d} = (x, y, z)$, where $x^2 + y^2 + z^2 = 1$. 

For numerical efficiency, we employ SH bases up to order $L=2$. The explicit forms for these basis functions, defined as polynomials of the vector components $(x, y, z)$, are given below:

\paragraph{Order $l=0$:}
\begin{equation}
    Y_0^0 = \frac{1}{2}\sqrt{\frac{1}{\pi}}
\end{equation}

\paragraph{Order $l=1$:}
\begin{align}
    Y_1^{-1} &= \frac{1}{2}\sqrt{\frac{3}{\pi}} y, \quad Y_1^{0} = \frac{1}{2}\sqrt{\frac{3}{\pi}} z, \quad Y_1^{1} = \frac{1}{2}\sqrt{\frac{3}{\pi}} x
\end{align}

\paragraph{Order $l=2$:}
\begin{align}
    Y_2^{-2} &= \frac{1}{2}\sqrt{\frac{15}{\pi}} xy \\
    Y_2^{-1} &= \frac{1}{2}\sqrt{\frac{15}{\pi}} yz \\
    Y_2^{0}  &= \frac{1}{4}\sqrt{\frac{5}{\pi}} (3z^2 - 1) \\
    Y_2^{1}  &= \frac{1}{2}\sqrt{\frac{15}{\pi}} xz \\
    Y_2^{2}  &= \frac{1}{4}\sqrt{\frac{15}{\pi}} (x^2 - y^2)
\end{align}

Note that these expressions are defined on the surface of the unit sphere. In implementation, $\mathbf{d}$ is computed as $\mathbf{d} = (\mathbf{o} - \bm{\mu}) / \|\mathbf{o} - \bm{\mu}\|_2$, where $\mathbf{o}$ is the camera optical center and $\bm{\mu}$ is the Gaussian mean position.

\section{Problem reformulation via geometric information}
\label{sec:prob_formulation}

To systematically address the failure modes of 4D scene reconstruction under trajectory constraints, we reframe the task from the lens of statistical parameter inference and geometric information theory. This allows us to quantify the uncertainty and identifiability of the scene parameters through the lens of information theory.

\subsection{Derivation of Fisher Information Matrix}

\paragraph{Likelihood and Log-Likelihood} 
We represent the persistent 4D scene by the parameter set $\theta = [\mathbf{g}^\top, \mathbf{s}^\top, \tau^\top]^\top$. The rendering process under a query context $\mathbf{z}$ synthesizes the pixel color $\mathbf{C}(\mathbf{z}; \theta)$ as the expectation of radiance along the camera ray: 
\begin{equation}
  \mathbf{C}(\mathbf{z}; \theta) = \sum_{k=1}^{M} \omega_k(\mathbf{g}_k, \mathbf{z}) \cdot c_k(s_{lm}^{(k)},  \tau_n^{(k)}, t, \mathbf{d})
\end{equation}

In a real-world sensing pipeline, the observed color $\mathbf{C}_{\text{obs}}$ is subject to measurement noise, which we model as additive Gaussian noise $\epsilon \sim \mathcal{N}(0, \sigma^2 \mathbf{I})$:
\begin{equation}
    \mathbf{C}_{\text{obs}} = \mathbf{C}(\mathbf{z}; \theta) + \epsilon.
\end{equation}
Under the assumption of independent and identically distributed (i.i.d.) noise across pixels, the likelihood of observing $\mathbf{C}_{\text{obs}}$ given the parameters $\theta$ is:
\begin{equation}
    p(\mathbf{C}_{\text{obs}} | \theta) = \frac{1}{(2\pi\sigma^2)^{N/2}} \exp \left( -\frac{\| \mathbf{C}_{\text{obs}} - \mathbf{C}(\mathbf{z}; \theta) \|^2}{2\sigma^2} \right).
\end{equation}
Maximizing this likelihood is equivalent to minimizing the negative log-likelihood $\mathcal{L}(\theta) = -\ln p(\mathbf{C}_{\text{obs}} | \theta)$. Ignoring constant terms, the objective becomes the standard Mean Squared Error (MSE):
\begin{equation}
    \mathcal{L}(\theta) = \frac{1}{2\sigma^2} \| \mathbf{C}_{\text{obs}} - \mathbf{C}(\mathbf{z}; \theta) \|^2.
\end{equation}
This formulation demonstrates that the MSE loss used in 4DGS optimization is statistically equivalent to the negative log-likelihood under a Gaussian noise model.

\paragraph{The score function and Fisher Information} 
The sensitivity of the likelihood to the parameters $\theta$ is captured by the \textbf{score function}, defined as the gradient of the log-likelihood:
\begin{equation}
    s(\theta) = \nabla_\theta \ln p(\mathbf{C}_{\text{obs}} | \theta) = \frac{1}{\sigma^2} \mathbf{J}^\top (\mathbf{C}_{\text{obs}} - \mathbf{C}(\mathbf{z}; \theta)),
\end{equation}
where $\mathbf{J} = \frac{\partial \mathbf{C}(\mathbf{z}; \theta)}{\partial \theta}$ is the spatiotemporal Jacobian matrix. The \textbf{Fisher Information Matrix (FIM)}, $F(\theta)$, is defined as the covariance of the score function:
\begin{equation}
    F(\theta) = \mathbb{E} [s(\theta) s(\theta)^\top].
\end{equation}
Substituting the score function into the above and noting that $\mathbb{E}[ \epsilon \epsilon^\top ] = \sigma^2 \mathbf{I}$, we derive the FIM for the 4D reconstruction task:
\begin{equation}
    F(\theta) = \frac{1}{\sigma^2} \mathbf{J}^\top \mathbf{J}.
\end{equation}

\paragraph{Statistical interpretation} 
The FIM $F(\theta)$ serves as a fundamental measure of \textbf{identifiability}. It characterizes the curvature of the information manifold; a high Fisher information along a certain parameter direction implies that the observation is highly sensitive to changes in that parameter, allowing for precise estimation. Conversely, an ill-conditioned or singular FIM indicates a lack of identifiability, where the observation fails to provide sufficient information to distinguish between different parameter configurations.

\subsection{Derivation of the Cramér-Rao Bound}

\paragraph{The "True Scene Parameter" assumption.} 
A fundamental prerequisite for deriving the standard CRB is the existence of a deterministic true parameter $\theta^*$. In the context of 4D scene reconstruction, we assume that the 4DGS representation possesses sufficient capacity to perfectly model the underlying continuous radiance field. Consequently, we assume there exists a set of true parameters $\theta^*$ such that rendering from any arbitrary continuous viewpoint yields the exact ground-truth image, subject only to the inherent observation noise $\epsilon$.

This physically intuitive assumption mathematically validates the concept of an "unbiased estimator" (i.e., the expectation of the estimated parameters equals this perfect configuration $\theta^*$) and allows us to perform rigorous local identifiability analysis. In the following derivation, $\theta$ strictly refers to this deterministic true parameter.

The Cramér-Rao Bound establishes the theoretical lower limit for the variance of any unbiased estimator. Let $\hat{\theta}(\mathbf{C}_{\text{obs}})$ be an unbiased estimator of the true parameters $\theta$, which by definition satisfies:
\begin{equation}
\label{eq:unbiased}
    \mathbb{E}[\hat{\theta}] = \int \hat{\theta} \, p(\mathbf{C}_{\text{obs}} | \theta) \, d\mathbf{C}_{\text{obs}} = \theta.
\end{equation}

Assuming standard regularity conditions that allow the interchange of differentiation and integration, we take the gradient of both sides of Eq.~\ref{eq:unbiased} with respect to $\theta$:
\begin{equation}
    \int \hat{\theta} \, \nabla_\theta p(\mathbf{C}_{\text{obs}} | \theta)^\top \, d\mathbf{C}_{\text{obs}} = \mathbf{I},
\end{equation}
where $\mathbf{I}$ is the identity matrix. Using the log-derivative trick, $\nabla_\theta p = p \nabla_\theta \ln p = p \cdot s(\theta)$, we can rewrite the integral as an expectation involving the score function:
\begin{equation}
\label{eq:cross_cov_1}
    \mathbb{E}[\hat{\theta} s(\theta)^\top] = \mathbf{I}.
\end{equation}

Furthermore, the expected value of the score function is strictly zero:
\begin{equation}
    \mathbb{E}[s(\theta)] = \int \nabla_\theta p(\mathbf{C}_{\text{obs}} | \theta) \, d\mathbf{C}_{\text{obs}} = \nabla_\theta \int p(\mathbf{C}_{\text{obs}} | \theta) \, d\mathbf{C}_{\text{obs}} = \nabla_\theta (1) = \mathbf{0}.
\end{equation}
Since $\mathbb{E}[s(\theta)] = \mathbf{0}$, we have $\mathbb{E}[\theta s(\theta)^\top] = \theta \mathbb{E}[s(\theta)]^\top = \mathbf{0}$. Subtracting this from Eq.~\ref{eq:cross_cov_1} yields the cross-covariance between the estimation error and the score function:
\begin{equation}
\label{eq:cross_cov_final}
    \mathbb{E}[(\hat{\theta} - \theta) s(\theta)^\top] = \mathbf{I}.
\end{equation}

Now, consider the joint random vector formed by the estimation error and the score function: $\mathbf{v} = \begin{bmatrix} \hat{\theta} - \theta \\ s(\theta) \end{bmatrix}$. Its block covariance matrix $\mathbf{\Sigma}$ must be positive semi-definite ($\succeq 0$):
\begin{equation}
    \mathbf{\Sigma} = \begin{bmatrix} 
    \mathbb{E}[(\hat{\theta} - \theta)(\hat{\theta} - \theta)^\top] & \mathbb{E}[(\hat{\theta} - \theta)s(\theta)^\top] \\ 
    \mathbb{E}[s(\theta)(\hat{\theta} - \theta)^\top] & \mathbb{E}[s(\theta)s(\theta)^\top] 
    \end{bmatrix} \succeq 0.
\end{equation}

Substituting the known terms—where $\mathbb{E}[(\hat{\theta} - \theta)(\hat{\theta} - \theta)^\top] = \text{Cov}(\hat{\theta})$, the cross-covariance is $\mathbf{I}$ (from Eq.~\ref{eq:cross_cov_final}), and the covariance of the score is the FIM $F(\theta)$—we obtain:
\begin{equation}
    \mathbf{\Sigma} = \begin{bmatrix} 
    \text{Cov}(\hat{\theta}) & \mathbf{I} \\ 
    \mathbf{I} & F(\theta) 
    \end{bmatrix} \succeq 0.
\end{equation}

According to the properties of block matrices, $\mathbf{\Sigma} \succeq 0$ implies that its Schur complement with respect to $F(\theta)$ must also be positive semi-definite. Assuming $F(\theta)$ is invertible, we have:
\begin{equation}
    \text{Cov}(\hat{\theta}) - \mathbf{I} F(\theta)^{-1} \mathbf{I}^\top \succeq 0 \quad \implies \quad \text{Cov}(\hat{\theta}) \succeq F(\theta)^{-1}.
\end{equation}

\textbf{Connection to identifiability:} This rigorous bound proves that the estimation variance is fundamentally bottlenecked by the inverse of the FIM. As analyzed in the main text, if the Jacobian $\mathbf{J}$ becomes rank-deficient due to trajectory coupling or transient visibility, $F(\theta)$ becomes singular. In this degenerate state, $F(\theta)^{-1} \to \infty$, mathematically guaranteeing the variance explosion and the subsequent catastrophic collapse during novel-view extrapolation.

\subsection{Diagnosing degeneracy via the Cramér-Rao Bound}
\label{sec:diag_degeneracy}

Having established that the FIM $F(\theta)$ governs the stability of parameter estimation, we now analyze its behavior under the specific constraints of autonomous driving trajectories. In these scenarios, the camera viewing direction $\mathbf{d}$ is typically a deterministic function of time $t$ along a 1D ego-trajectory $\gamma(t)$, i.e., $\mathbf{d} = \gamma(t)$. 

\textbf{Subspace inclusion and rank deficiency.} 
To analyze the entanglement between spatial and temporal attributes, we partition the joint Jacobian $\mathbf{J} = [\mathbf{J}_s, \mathbf{J}_\tau]$, where $\mathbf{J}_s$ and $\mathbf{J}_\tau$ represent the partial derivatives with respect to spatial and temporal parameters, respectively. The resulting joint FIM is given by:
\begin{equation}
    F(\theta) = \frac{1}{\sigma^2} \begin{bmatrix} \mathbf{J}_s^\top \mathbf{J}_s & \mathbf{J}_s^\top \mathbf{J}_\tau \\ \mathbf{J}_\tau^\top \mathbf{J}_s & \mathbf{J}_\tau^\top \mathbf{J}_\tau \end{bmatrix}.
\end{equation}
In an ideal multi-view concurrent sampling scenario, $\mathbf{J}_s$ and $\mathbf{J}_\tau$ are mutually orthogonal, leading to a well-conditioned block-diagonal FIM. However, on a smooth 1D trajectory, the spatial basis functions transform into temporal signals. As the temporal basis set $\{\phi_n(t)\}$ is dense in the Hilbert space $L^2([0, T])$, any trajectory-projected spatial signal can be approximated by the temporal manifold with arbitrary precision. This leads to a \textbf{subspace inclusion} $\mathcal{R}(\mathbf{J}_s) \subseteq \mathcal{R}(\mathbf{J}_\tau)$, rendering the joint Jacobian rank-deficient and the FIM singular.

\subsection{Information-theoretic analysis of identifiability failure}

To rigorously analyze the identifiability failure in 4D scene reconstruction under 1D trajectories, we employ the framework of the Fisher Information Matrix (FIM) and the Cramér-Rao Bound (CRB). We consider the joint parameter space $\Theta = [\mathbf{s}^\top, \bm{\tau}^\top]^\top$, where $\mathbf{s}$ and $\bm{\tau}$ denote the spatial (view-dependent) and temporal parameters, respectively. The FIM $\mathbf{F}$ can be partitioned into a block structure:
\begin{equation}
\mathbf{F} = \begin{bmatrix} 
\mathbf{F}_{ss} & \mathbf{F}_{s\tau} \\ 
\mathbf{F}_{\tau s} & \mathbf{F}_{\tau\tau} 
\end{bmatrix}
\end{equation}
According to the Cramér-Rao Bound, the covariance lower bound for any unbiased estimator of the spatial parameters $\mathbf{s}$ is given by the corresponding block of the inverse FIM. Using the matrix inversion lemma, we have:
\begin{equation}
\text{Cov}(\hat{\mathbf{s}}) \succeq [\mathbf{F}^{-1}]_{ss} = (\mathbf{F}_{ss} - \mathbf{F}_{s\tau} \mathbf{F}_{\tau\tau}^{-1} \mathbf{F}_{\tau s})^{-1}
\end{equation}
The term $\mathbf{S} = \mathbf{F}_{ss} - \mathbf{F}_{s\tau} \mathbf{F}_{\tau\tau}^{-1} \mathbf{F}_{\tau s}$ is the Schur complement of $\mathbf{F}_{\tau\tau}$ in $\mathbf{F}$. In the context of information theory, $\mathbf{S}$ represents the effective Fisher information—the residual information available for estimating $\mathbf{s}$ after marginalizing out the influence of the temporal parameters $\bm{\tau}$.

We now demonstrate how the trajectory-induced coupling leads to the collapse of $\mathbf{S}$. Let $\mathbf{J}_s$ and $\mathbf{J}_\tau$ be the partial Jacobians with respect to $\mathbf{s}$ and $\bm{\tau}$. The FIM blocks are defined as:
\begin{equation}
\mathbf{F}_{ss} = \mathbf{J}_s^\top \mathbf{J}_s, \quad \mathbf{F}_{\tau\tau} = \mathbf{J}_\tau^\top \mathbf{J}_\tau, \quad \mathbf{F}_{s\tau} = \mathbf{J}_s^\top \mathbf{J}_\tau
\end{equation}
In a restricted 1D ego-trajectory $\gamma(t)$, any viewpoint-induced spatial variation $\mathbf{J}_s$ can be approximated by the expressive temporal manifold $\mathbf{J}_\tau$ due to the deterministic mapping $\mathbf{d} = \gamma(t)$. This implies the existence of an attribution matrix $\mathbf{A}$ such that:
\begin{equation}
\mathbf{J}_s = \mathbf{J}_\tau \mathbf{A}
\end{equation}
Substituting this relation into the FIM components, we obtain:
\begin{align}
\mathbf{F}_{ss} &= (\mathbf{J}_\tau \mathbf{A})^\top (\mathbf{J}_\tau \mathbf{A}) = \mathbf{A}^\top \mathbf{F}_{\tau\tau} \mathbf{A} \\
\mathbf{F}_{s\tau} &= (\mathbf{J}_\tau \mathbf{A})^\top \mathbf{J}_\tau = \mathbf{A}^\top \mathbf{F}_{\tau\tau} \\
\mathbf{F}_{\tau s} &= \mathbf{J}_\tau^\top (\mathbf{J}_\tau \mathbf{A}) = \mathbf{F}_{\tau\tau} \mathbf{A}
\end{align}
Evaluating the Schur complement under this condition yields:
\begin{equation}
\mathbf{S}_{s} = \mathbf{A}^\top \mathbf{F}_{\tau\tau} \mathbf{A} - (\mathbf{A}^\top \mathbf{F}_{\tau\tau}) \mathbf{F}_{\tau\tau}^{-1} (\mathbf{F}_{\tau\tau} \mathbf{A}) = \mathbf{0}
\end{equation}
The collapse of the effective Fisher information ($\mathbf{S} \to \mathbf{0}$) causes the variance lower bound to diverge to infinity ($\text{Cov}(\hat{\mathbf{s}}) \to \infty$). Mathematically, this confirms that the spatial manifold becomes statistically unidentifiable along the 1D training corridor.

To understand why the temporal parameters dominate the learning process, we further analyze the estimation variance of $\bm{\tau}$. Symmetrically, the CRB for the temporal parameters is bounded by the inverse of the Schur complement of $\mathbf{F}_{ss}$:
\begin{equation}
\text{Cov}(\hat{\bm{\tau}}) \succeq (\mathbf{F}_{\tau\tau} - \mathbf{F}_{\tau s} \mathbf{F}_{ss}^{-1} \mathbf{F}_{s\tau})^{-1}
\end{equation}
Let $\mathbf{S}_\tau = \mathbf{F}_{\tau\tau} - \mathbf{F}_{\tau s} \mathbf{F}_{ss}^{-1} \mathbf{F}_{s\tau}$ denote the effective Fisher information for $\bm{\tau}$. Substituting $\mathbf{J}_s = \mathbf{J}_\tau \mathbf{A}$ into $\mathbf{S}_\tau$ yields:
\begin{equation}
\mathbf{S}_\tau = \mathbf{F}_{\tau\tau} - \mathbf{F}_{\tau\tau} \mathbf{A} ( \mathbf{A}^\top \mathbf{F}_{\tau\tau} \mathbf{A} )^{-1} \mathbf{A}^\top \mathbf{F}_{\tau\tau}
\end{equation}
Here, the subtracted term acts as an oblique projection operator onto the subspace spanned by the spatial basis. 

An essential asymmetry arises here: since the temporal manifold is typically designed with a much higher dimensionality than the spatial view-dependency to capture complex motions, the matrix $\mathbf{A}$ is a tall matrix. Consequently, while $\mathbf{S}_{s}$ for the spatial parameters completely collapses to $\mathbf{0}$, $\mathbf{S}_\tau$ does not vanish but only suffers a rank deficiency. $\bm{\tau}$ retains non-zero effective information in the directions orthogonal to the spatial manifold. This mathematical asymmetry provides the exact mechanism for gradient hijacking: optimization algorithms naturally gravitate towards updating $\bm{\tau}$ due to its non-zero information gradients, leaving $\mathbf{s}$ hopelessly under-optimized.

\section{Proof of identifiability restoration via OPG}
\label{sec:appendix_opg_proof}

In this section, we provide the detailed derivation of the Cramér-Rao Bounds (CRB) under the Orthogonal Projected Gradient (OPG) scheme. 

\paragraph{1. Orthogonality of the purified Jacobian.} 
The OPG scheme defines the purified temporal Jacobian as $\tilde{\mathbf{J}}_\tau = \mathbf{P}_s^\perp \mathbf{J}_\tau$, where $\mathbf{P}_s^\perp = \mathbf{I} - \mathbf{J}_s (\mathbf{J}_s^\top \mathbf{J}_s)^{-1} \mathbf{J}_s^\top$ is the projector onto the null-space of $\mathbf{J}_s$. We first show that $\mathbf{J}_s$ and $\tilde{\mathbf{J}}_\tau$ are strictly orthogonal:
\begin{align}
    \mathbf{J}_s^\top \tilde{\mathbf{J}}_\tau &= \mathbf{J}_s^\top (\mathbf{I} - \mathbf{J}_s (\mathbf{J}_s^\top \mathbf{J}_s)^{-1} \mathbf{J}_s^\top) \mathbf{J}_\tau \\
    &= (\mathbf{J}_s^\top - \mathbf{J}_s^\top \mathbf{J}_s (\mathbf{J}_s^\top \mathbf{J}_s)^{-1} \mathbf{J}_s^\top) \mathbf{J}_\tau \\
    &= (\mathbf{J}_s^\top - \mathbf{J}_s^\top) \mathbf{J}_\tau = \mathbf{0}
\end{align}
This proves that the cross-modal Fisher information blocks $\mathbf{F}_{s\tilde{\tau}}$ and $\mathbf{F}_{\tilde{\tau}s}$ are zero matrices.

\paragraph{2. Block-diagonalization of the FIM.}
The joint FIM under OPG is constructed as:
\begin{equation}
    \mathbf{F}_{OPG} = \frac{1}{\sigma^2} \begin{bmatrix} \mathbf{J}_s \\ \tilde{\mathbf{J}}_\tau \end{bmatrix}^\top \begin{bmatrix} \mathbf{J}_s & \tilde{\mathbf{J}}_\tau \end{bmatrix} = \frac{1}{\sigma^2} \begin{bmatrix} \mathbf{J}_s^\top \mathbf{J}_s & \mathbf{J}_s^\top \tilde{\mathbf{J}}_\tau \\ \tilde{\mathbf{J}}_\tau^\top \mathbf{J}_s & \tilde{\mathbf{J}}_\tau^\top \tilde{\mathbf{J}}_\tau \end{bmatrix}
\end{equation}
Substituting the orthogonality result $\mathbf{J}_s^\top \tilde{\mathbf{J}}_\tau = \mathbf{0}$, we obtain a block-diagonal matrix:
\begin{equation}
    \mathbf{F}_{OPG} = \frac{1}{\sigma^2} \begin{bmatrix} \mathbf{J}_s^\top \mathbf{J}_s & \mathbf{0} \\ \mathbf{0} & \tilde{\mathbf{J}}_\tau^\top \tilde{\mathbf{J}}_\tau \end{bmatrix} = \begin{bmatrix} \mathbf{F}_{ss} & \mathbf{0} \\ \mathbf{0} & \mathbf{F}_{\tilde{\tau}\tilde{\tau}} \end{bmatrix}
\end{equation}

\paragraph{3. Derivation of the decoupled CRB and temporal variance.}
For a block-diagonal FIM, the Schur complements simplify significantly. The effective information for the spatial parameters $\mathbf{s}$ is:
\begin{equation}
    \mathbf{S}_s = \mathbf{F}_{ss} - \mathbf{F}_{s\tilde{\tau}} \mathbf{F}_{\tilde{\tau}\tilde{\tau}}^{-1} \mathbf{F}_{\tilde{\tau}s} = \mathbf{F}_{ss} - \mathbf{0} = \mathbf{F}_{ss}.
\end{equation}
The resulting lower bound for the estimation covariance is strictly bounded:
\begin{equation}
    \text{Cov}(\hat{\mathbf{s}}) \succeq \mathbf{S}_s^{-1} = \sigma^2 (\mathbf{J}_s^\top \mathbf{J}_s)^{-1}.
\end{equation}
This confirms that OPG successfully restores the identifiability of spatial parameters by isolating them from temporal interference.

Similarly, the effective information for the temporal parameters $\tau$ is derived as:
\begin{equation}
    \mathbf{S}_{\tau} = \mathbf{F}_{\tilde{\tau}\tilde{\tau}} - \mathbf{F}_{\tilde{\tau}s} \mathbf{F}_{ss}^{-1} \mathbf{F}_{s\tilde{\tau}} = \mathbf{F}_{\tilde{\tau}\tilde{\tau}} = \frac{1}{\sigma^2} \tilde{\mathbf{J}}_\tau^\top \tilde{\mathbf{J}}_\tau.
\end{equation}
However, because $\tilde{\mathbf{J}}_\tau = \mathbf{P}_s^\perp \mathbf{J}_\tau$ is a projected operator, it explicitly removes all temporal gradient components that overlap with the spatial manifold. Consequently, $\mathbf{S}_{\tau}$ is strictly rank-deficient (singular). For singular information matrices, the CRB is evaluated using the Moore-Penrose pseudoinverse $(\cdot)^\dagger$:
\begin{equation}
    \text{Cov}(\hat{\tau}) \succeq \mathbf{S}_{\tau}^{\dagger} = \sigma^2 (\tilde{\mathbf{J}}_\tau^\top \tilde{\mathbf{J}}_\tau)^{\dagger}.
\end{equation}

\textbf{Physical implication (The cost of OPG):} While $\text{Cov}(\hat{\tau})$ is bounded within the purified subspace (the non-zero eigenvalues of $\mathbf{S}_{\tau}$), the estimation variance of $\tau$ along the projected null-space approaches \textbf{infinity}. This means that while OPG saves the spatial parameters, it leaves the temporal parameters highly vulnerable and unconstrained in certain directions. This mathematical vulnerability perfectly necessitates the introduction of the Temporal Total Variation (TV) prior to bound the remaining infinite variance of $\tau$.

\section{Inplementation Details}
\label{appendix:inplementation}
All models are trained on a single NVIDIA A100 GPU at a resolution of $1600 \times 1066$ with three front cameras. Following the 3:1 interpolation protocol \cite{yan2024street}, we withhold every fourth frame for testing while using the remaining three for training. To ensure a fair comparison, all methods are evaluated on the full sequence of each case (approximately 200 frames) and initialized using LiDAR point clouds. All models are trained for 30,000 iterations except for $S^3$Gaussian, which follows its original configuration. We set fourier dimension to 16 for all methods related with fourier basis, accordingly the control points of B-spline is also set to 16. SH degree is set as 1. The TV regularization weight is set to $\lambda_{TV} = 0.005$. We employ a \textbf{hierarchical training scheme} via OPG. We \textbf{freeze the time-varying parameters} and perform standard optimization for the first 23,000 iterations, then perform \textbf{OPG} once, after which the \textbf{geometric and view-dependent parameters are frozen} to allow for the optimization of \textbf{temporal parameters}. The integration of OPG scheme and TV regularization introduces negligible computational overhead, maintaining a training time almost identical to the original baseline.

\section{Licenses and URLs of existing assets}
\label{sec:licenses}
In this work, we utilize several existing open-source assets for our experiments. We strictly respect their licenses and terms of use. The details are as follows:

\begin{itemize}
    \item \textbf{Waymo Open Dataset (v1.4.1)} \cite{sun2020scalability}: Available at \url{https://waymo.com/open/}. Used under the Waymo Dataset License Agreement, strictly for non-commercial academic research purposes.
    \item \textbf{StreetGaussians} \cite{yan2024street}: The official codebase (\url{https://github.com/zju3dv/street_gaussians}) is utilized under its custom academic license. We confirm that our usage is strictly for educational, research, and non-profit baseline evaluation purposes, in full compliance with its non-commercial terms.
    \item \textbf{$S^3$Gaussian} \cite{huang2024s3}: The official codebase (\url{https://github.com/nnanhuang/S3Gaussian}) is utilized under the Gaussian-Splatting License (Inria and MPII). We confirm that our usage is strictly for non-commercial research and evaluation purposes, in full compliance with its terms.
\end{itemize}

\end{document}